\documentclass[acmsmall,screen]{acmart}
\AtBeginDocument{%
  \providecommand\BibTeX{{%
    \normalfont B\kern-0.5em{\scshape i\kern-0.25em b}\kern-0.8em\TeX}}}
\pdfoutput=1 
  
\setcopyright{acmcopyright}
\copyrightyear{20xx}
\acmYear{20xx}

\acmJournal{CSUR}

\renewcommand\footnotetextcopyrightpermission[1]{} 
\setcopyright{none}

\usepackage{chngcntr}

\usepackage{tocloft}
\usepackage{graphicx} %

\usepackage{caption} %
\usepackage{algorithm}
\usepackage{listings}
\usepackage{algorithmic}
\usepackage{amsmath}
\usepackage{booktabs}
\usepackage{multirow}
\usepackage[outdir=./]{epstopdf}
\usepackage{enumitem}
\usepackage{caption}
\usepackage{subcaption}
\usepackage{subfloat}
\usepackage{newfloat}
\usepackage{svg} %
\usepackage[normalem]{ulem}
\usepackage{framed}
\usepackage{mdframed}
\usepackage{xcolor}
\usepackage{lipsum}
\usepackage{float}
\usepackage{hyperref}
\usepackage{hyperxmp}
\usepackage{url}
\usepackage{amsfonts}
\usepackage{makecell}
\usepackage{array}
\usepackage{multirow}

\definecolor{shadecolor}{gray}{0.9}

\newlist{todolist}{itemize}{2}
\setlist[todolist]{label=$\square$}
\usepackage{pifont}

\usepackage{array}

\begin{document}
\title{Understanding World or Predicting Future? A Comprehensive Survey of World Models}

\author{Jingtao~Ding$^{*}$, Yunke~Zhang$^{*}$, Yu~Shang$^{*}$, Jie~Feng$^{*}$, Yuheng~Zhang$^{\dag}$, Zefang~Zong$^{\dag}$, Yuan~Yuan$^{\dag}$, Hongyuan~Su$^{\dag}$, Nian~Li$^{\dag}$, Jinghua~Piao$^{\dag}$, Yucheng~Deng, Nicholas~Sukiennik, Chen~Gao, Fengli~Xu, Yong~Li}
\email{dingjt15@tsinghua.org.cn, liyong07@tsinghua.edu.cn}
\thanks{*These authors contributed equally.}
\thanks{{\dag}These authors contributed equally.}
\affiliation{%
  \institution{Department of Electronic Engineering, Beijing National Research Center for Information Science and Technology (BNRist), Tsinghua University}
  \country{China}
  }

\renewcommand{\shortauthors}{Ding and Zhang et al.}

\begin{abstract}
The concept of world models has garnered significant attention due to advancements in multimodal large language models such as GPT-4 and video generation models such as Sora, which are central to the pursuit of artificial general intelligence. This survey offers a comprehensive review of the literature on world models. Generally, world models are regarded as tools for either understanding the present state of the world or predicting its future dynamics. This review presents a systematic categorization of world models, emphasizing two primary functions: (1) constructing internal representations to understand the mechanisms of the world, and (2) predicting future states to simulate and guide decision-making. Initially, we examine the current progress in these two categories. We then explore the application of world models in key domains, including generative games, autonomous driving, robotics, and social simulacra, with a focus on how each domain utilizes these aspects. Finally, we outline key challenges and provide insights into potential future research directions. We summarize the representative papers along with their code repositories in \url{https://github.com/tsinghua-fib-lab/World-Model}.
\end{abstract}

\maketitle

\section{Introduction}

The scientific community has long aspired to develop a unified model that can replicate its fundamental dynamics of the world in pursuit of Artificial General Intelligence (AGI)~\cite{lecun2022path}. In 2024, the emergence of multimodal large language models (LLMs) and video generation models like Sora~\cite{sora2024} has intensified discussions surrounding such \textbf{World Models}. While these models demonstrate an emerging capacity to capture aspects of world knowledge--such as Sora's generated videos, which appear to perfectly adhere to physical laws--questions persist regarding whether they truly qualify as comprehensive world models. Therefore, a systematic review of recent advancements, applications, and future directions in world model research is both timely and essential as we look toward new breakthroughs in the era of artificial intelligence (AI).

The definition of a world model remains a subject of ongoing debate, generally divided into two primary perspectives: \textit{understanding the world} and \textit{predicting the future}. As depicted in Figure~\ref{fig:intro}, early work by Ha and Schmidhuber~\cite{ha2018world} focused on abstracting the external world to gain a deep understanding of its underlying mechanisms. In contrast, LeCun~\cite{lecun2022path} argued that a world model should not only perceive and model the real world but also possess the capacity to envision possible future states to inform decision-making. Video generation models such as Sora represent an approach that concentrates on simulating future world evolution and thus align more closely with the predictive aspect of world models. This raises the question of whether a world model should prioritize understanding the present or forecasting future states. In this paper, we provide a comprehensive review of the literature from both perspectives, highlighting key approaches and challenges.

The potential applications of world models span a wide array of fields, each with distinct requirements for understanding and predictive capabilities. In autonomous driving, for example, world models need to perceive road conditions in real-time~\cite{yolop_2022, teichmann2018multinetrealtimejointsemantic} and accurately predict their evolution~\cite{ngiam2021scene, shi2022motion, zhou2023query}, with a particular focus on immediate environmental awareness and forecasting of complex trends. For robotics, world models are essential for tasks such as navigation~\cite{shah2023gnm}, object detection~\cite{wang2024repvit}, and task planning~\cite{hafner2019learning}, requiring a precise understanding of external dynamics~\cite{gao2023s} and the ability to generate interactive and embodied environments~\cite{park2023generative}. In the realm of simulation of virtual social systems, world models must capture and predict more abstract behavioral dynamics, such as social interactions and human decision-making processes. Thus, a comprehensive review of advancements in these capabilities, alongside an exploration of future research directions and trends, is both timely and essential.

Existing surveys on world models can generally be classified into two categories, as shown in Table~\ref{tab:survey_compare}. The first category primarily focuses on describing the application of world models in specific fields such as video processing and generation~\cite{cho2024sora, zhu2024sora}, autonomous driving~\cite{guan2024world, li2024data, yan2024forging}, and agent-based applications~\cite{zhu2024sora}. The second category~\cite{mai2024efficient} concentrates on the technological transitions from multi-modal models, which are capable of processing data across various modalities, to world models. However, these papers often lack a systematic examination of what precisely constitutes a world model and what different real-world applications require from these models. In this article, we aim to formally define and categorize world models, review recent technical progress, and explore their extensive applications.

The main contributions of this survey can be summarized as follows: (1) We present a novel categorization system for world models structured around two primary functions: \textit{constructing implicit representations to understand the mechanism of the external world} and \textit{predicting future states of the external world}. The first category focuses on the development of models that learn and internalize world knowledge to support subsequent decision-making, while the latter emphasizes enhancing predictive and simulative capabilities in the physical world from visual perceptions. (2) Based on this categorization, we classify how various key application areas, including generative games, autonomous driving, robots, and social simulacra, emphasize different aspects of world models. (3) We highlight future research directions and trends of world models that can adapt to a broader spectrum of practical applications.

The remainder of this paper is organized as follows. In Section~\ref{sec:background}, we introduce the background of the world model and propose our categorization system. Section~\ref{sec::inner} and Section~\ref{sec::future} elaborate on the details of current research progress on two categories of world models, respectively. Section~\ref{sec:application} covers applications of the world model in three key research fields. Section~\ref{sec:discussion} outlines open problems and future directions of world models.

\section{Background and Categorization} \label{sec:background}

\begin{figure}[t]
    \centering
    \includegraphics[width=1.0\linewidth]{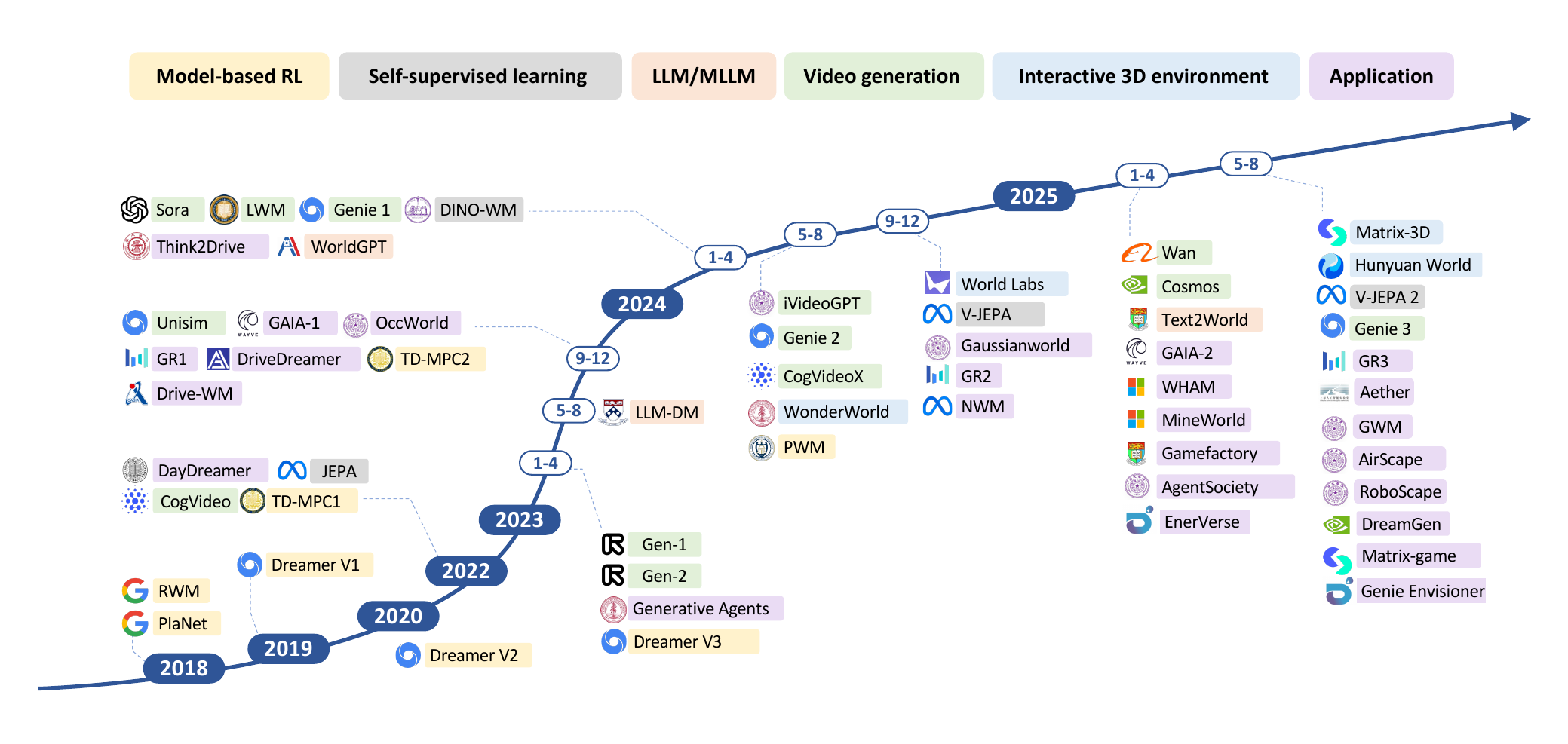}
    \caption{The roadmap of world models in deep learning era. }
    \label{fig:intro}
\end{figure}

\subsection{History and Current Development}
In this section, we explore the evolving concepts of world models in the literature and categorize efforts to construct world models into two distinct branches: internal representation and future prediction.

\textbf{Pre Deep Learning Era.} The concept of building an internal model of the world has a long history in AI, dating back to foundational work such as Marvin Minsky's frame representation in the 1960s~\cite{minsky1974framework}, designed to systematically capture structured knowledge about the world. In the context of reinforcement learning, world models emerged as a fundamental component of model-based approaches, where agents construct explicit representations of their environment's dynamics. Early work in this domain focused on learning transition models that could predict the next state given the current state and action~\cite{sutton1990integrated}, enabling agents to perform planning and simulate potential action sequences before execution. These environment models were typically represented using tabular methods or simple parametric functions, laying the groundwork for more sophisticated world modeling approaches that would emerge with the advent of deep learning.

\textbf{Model-based Reinforcement Learning.} 
Ha \textit{et al.}~\cite{ha2018recurrent,ha2018world} significantly revived and popularized the term ``world model'' in 2018 by proposing a recurrent neural-network-based implicit model for learning latent representations. This line of research aligns with the psychological theory of ``mental models''~\cite{johnson1983mental}\footnote{\url{https://plato.stanford.edu/entries/mental-representation/}}, which posits that humans perceive the external world by abstracting it into simplified elements and relationships—an underlying philosophical principle reflected in both frames and world models.
This principle suggests that our understanding of the world, when viewed from a cognitive perspective, typically involves constructing abstract representations that capture essential patterns without requiring exhaustive detail. Building upon this conceptual framework, the authors introduce an agent module inspired by the human cognitive system, as illustrated in Figure~\ref{fig:intro}. In this recurrent world model~(RWM), the agent receives feedback from the real-world environment, which is then transformed into a series of inputs that train the model. This model is adept at simulating potential outcomes following specific actions within the external environment. Essentially, it creates a mental simulation of potential future world evolutions, with decisions made based on the predicted outcomes of these states. This methodology closely mirrors the model-based reinforcement learning method, where both strategies involve the model generating internal representations of the external world to facilitate navigation through and resolution of various decision-making tasks. 
Following this conceptual foundation, subsequent developments have further advanced world model architectures, including Google DeepMind's Dreamer series~\cite{hafner2019dream,hafner2020mastering,hafner2025mastering}, which has demonstrated the scalability and effectiveness of learned world representations across increasingly complex domains.

\textbf{Self-supervised Learning.} 
In the visionary article on the development of autonomous machine intelligence in 2022~\cite{lecun2022path}, Yann LeCun introduced the Joint Embedding Predictive Architecture (JEPA), a framework mirroring the human brain's structure. As illustrated in Figure~\ref{fig:intro}, JEPA comprises a perception module that processes sensory data~(i.e., an encoder), followed by a cognitive module~(i.e., a predictor) that evaluates this information, effectively embodying the world model. This model allows the brain to assess actions and determine the most suitable responses for real-world applications.
A key innovation of JEPA lies in its self-supervised learning paradigm, which enables the system to learn rich representations of the world without relying on extensive labeled data. Rather than predicting raw sensory inputs in pixel space, JEPA learns to predict abstract representations in a latent embedding space, making the learning process more efficient and robust. This approach allows the model to capture semantic relationships and causal structures in the data while avoiding the computational burden and potential pitfalls of pixel-level prediction, such as focusing on irrelevant details or noise.
LeCun's framework is particularly intriguing due to its incorporation of the dual-system concept, mirroring "fast" and "slow" thinking. System 1 involves intuitive, instinctive reactions: quick decisions made without explicit world model consultation, such as instinctively dodging an oncoming person. In contrast, System 2 employs deliberate, calculated reasoning that leverages the learned world model to consider future states. It extends beyond immediate sensory input, simulating potential future scenarios through the self-supervised representations, like predicting events in a room over the next ten minutes and adjusting actions accordingly. This level of foresight requires constructing a world model that can effectively guide decisions based on the anticipated dynamics and evolution of the environment.
In this framework, the world model is essential for understanding and representing the external world through self-supervised learning of latent variables, which capture key information while filtering out redundancies. This approach allows for a highly efficient, minimalistic representation of the world, facilitating optimal decision-making and planning for future scenarios. Building upon these principles, recent implementations such as V-JEPA~\cite{bardes2024revisiting} and V-JEPA2~\cite{assran2025v} have demonstrated the practical viability of video-based self-supervised learning, showing how JEPA architectures can learn rich spatiotemporal representations from unlabeled video data for downstream vision tasks.

\textbf{Large Language Models.} 
"The limits of my language mean the limits of my world."—Ludwig Wittgenstein. This profound observation finds particular relevance in the context of large language models, which learn fundamental principles of world operation through textual data that can be leveraged to construct comprehensive world models. Recent research has demonstrated that LLMs trained on vast corpora naturally acquire latent world knowledge, including spatial and temporal understanding, enabling them to make sophisticated predictions about real-world scenarios~\cite{gurnee2023language, manvi2023geollm}. This capability has been harnessed for model-based task planning, where pre-trained language models serve as the foundation for constructing world models that can reason about complex sequential tasks~\cite{guan2023leveraging}. The integration of multimodal capabilities further enhances world modeling potential. Multimodal Large Language Models (MLLMs) can process and integrate information across visual, textual, and other sensory modalities, creating richer and more comprehensive world representations~\cite{ge2024worldgpt}. 
Understanding how these models process, represent, and utilize world knowledge remains crucial for developing more effective world models that can bridge the gap between linguistic knowledge and real-world understanding~\cite{yang2025thinking}.

\textbf{Video Generation.} 
Video generation has emerged as the predominant approach to world modeling in contemporary AI research. Unlike earlier implicit world representations, these models explicitly generate visual sequences that demonstrate understanding of temporal dynamics, spatial consistency, and physical laws. Powered by advanced generative techniques such as diffusion modeling and transformer architectures, recent video generation models including Sora~\cite{sora2024}, Keling~\cite{keling2024}, and Gen-2~\cite{runway2023} take text instructions or real-world visual data as input and produce high-quality video sequences. These models demonstrate exceptional world modeling capabilities, including maintaining consistency in 3D video simulations, producing physically plausible outcomes, and simulating complex digital environments.
The sophistication of these approaches suggests they model underlying real-world dynamics rather than merely generating visually appealing content. This represents a fundamental shift toward world models that can actively simulate and predict how environments evolve over time. Recent developments have further advanced this paradigm, with Cosmos~\cite{agarwal2025cosmos} achieving breakthrough performance in physics law adherence and Genie 3~\cite{genie3} enabling real-time interaction capabilities for controllable world simulation.

\textbf{Interactive 3D Environments.} 
Interactive 3D scene generation represents another important paradigm in world modeling, focusing on creating immersive 3D worlds that enable spatial exploration and user interaction within virtual environments. Representative work such as Wonderworld demonstrates the capability to generate interactive 3D scenes from a single 2D image~\cite{yu2025wonderworld}, showcasing the potential for creating explorable virtual worlds from minimal input. This approach emphasizes spatial consistency, geometric understanding, and real-time responsiveness to user navigation and interaction. Recent advances have significantly expanded these capabilities, with Matrix-3D achieving wide-coverage omnidirectional explorable 3D world generation through panoramic 3D reconstruction~\cite{yang2025matrix}, and HunyuanWorld 1.0 enabling immersive 360° experiences through semantically layered 3D mesh representations that provide seamless compatibility with existing computer graphics pipelines~\cite{team2025hunyuanworld}.

\textbf{Applications.}
World models have rapidly expanded across diverse application domains since 2023. In autonomous driving, foundational works such as GAIA-1 and Drive-WM established approaches for modeling vehicle interactions and environmental dynamics in complex traffic scenarios~\cite{hu2023gaia,wang2023drivingfuturemultiviewvisual}. The robotics domain has similarly advanced, exemplified by DayDreamer in 2023~\cite{wu2023daydreamer} and continuing with recent developments in 2025 for robotic manipulation tasks~\cite{lu2025gwm}.
Navigation applications have emerged with robot path planning~\cite{bar2025navigation} extending to six-degree-of-freedom aerial agents~\cite{zhao2025airscape}. Gaming represents a particularly promising domain, with landmark work on world and human action models (WHAM) demonstrating how world models can create dynamic, responsive virtual environments~\cite{kanervisto2025world}. At the larger scale, agent-based social simulation leverages world models to understand complex societal dynamics and human interactions, offering computational insights into real-world social phenomena~\cite{piao2025agentsociety}.

\subsection{Evolving Concept from Multiple Domains}
The concept of world models in artificial intelligence has deep psychological roots that extend far beyond contemporary machine learning. Understanding these foundational connections reveals how modern AI world models represent a computational realization of fundamental cognitive principles that have been studied for decades across multiple disciplines.

The psychological concept of mental models was first articulated by Scottish psychologist Kenneth Craik in his seminal work "The Nature of Explanation" (1943), where he proposed that "the mind constructs small-scale models of reality" to predict and understand external events~\cite{craik1943nature}. Craik's insight was that human cognition fundamentally operates by creating internal representations that capture the essential structure and dynamics of the external world, enabling predictive reasoning and adaptive behavior.

This foundational concept was systematically developed and formalized by British psychologist Philip Nicholas Johnson-Laird in the 1980s through his Mental Models Theory. In his influential work "Mental Models: Towards a Cognitive Science of Language, Inference, and Consciousness" (1983), Johnson-Laird demonstrated that human reasoning operates through the construction and manipulation of mental models—internal representations that preserve the structural relationships of the situations they represent~\cite{johnson1983mental}. According to this theory, when humans engage in deductive reasoning, inductive inference, or counterfactual thinking, they mentally simulate different scenarios by constructing and testing alternative models of possible worlds.

Johnson-Laird's framework established several key principles that directly parallel contemporary AI world models: mental models are finite representations of potentially infinite domains, they capture structural relationships rather than superficial details, and they enable predictive simulation of alternative scenarios. These principles have become fundamental to understanding how both human and artificial agents can efficiently represent and reason about complex environments.

\begin{figure}[t]
    \centering
    \includegraphics[width=1.0\linewidth]{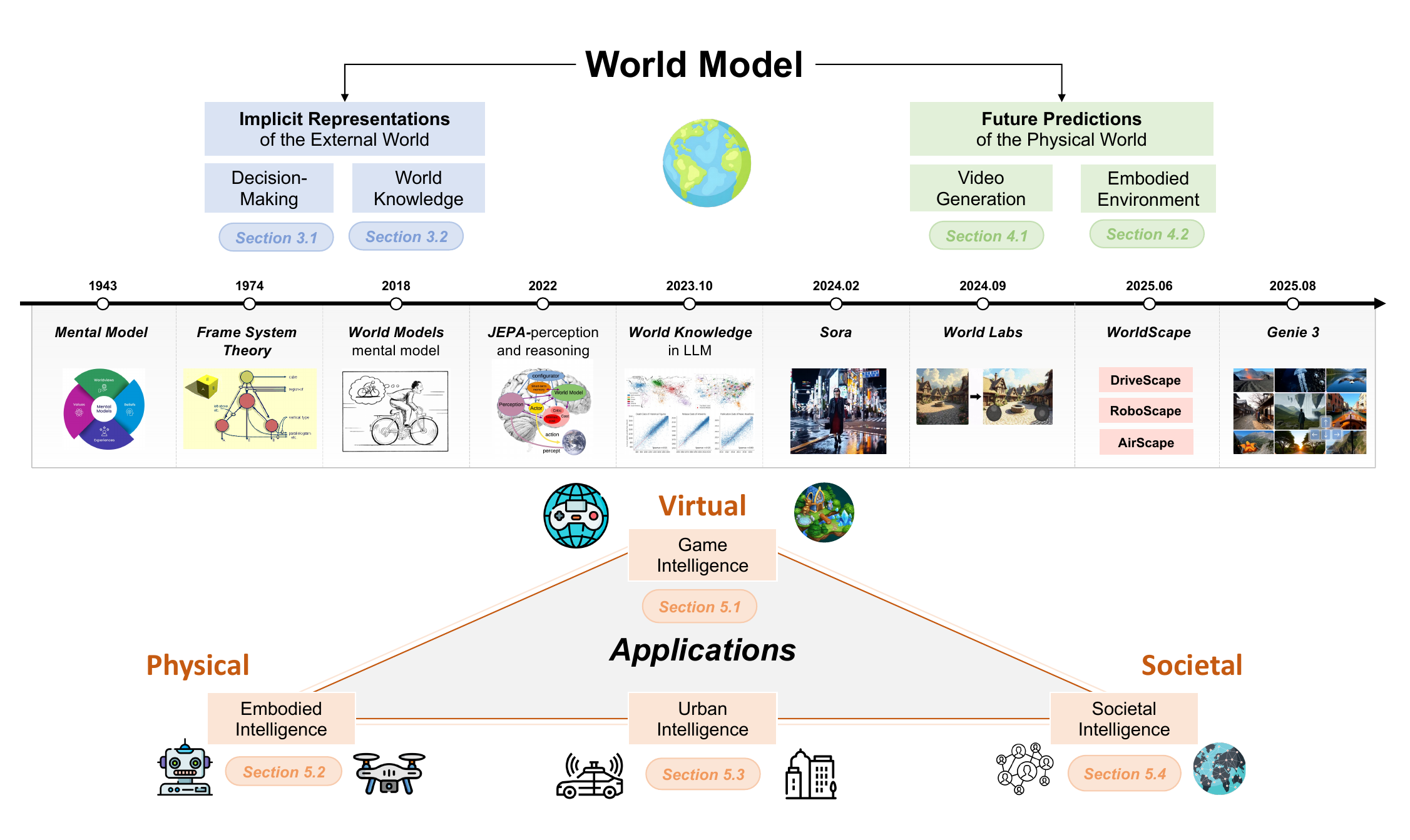}
    \caption{The overall framework of this survey. We systematically define the essential purpose of a world model as understanding the dynamics of the external world and predicting future scenarios. The timeline illustrates the development of key definitions and applications. }
    \label{fig:intro}
\end{figure}

\subsection{Categorization}
Whether focusing on learning internal representations of the external world or simulating its operational principles, these concepts coalesce into a shared consensus: the essential purpose of a world model is to understand the dynamics of the world and compute the next state with certainty (or with some guarantee), which empowers the model to extrapolate longer-horizon evolution and to support downstream decision-making and planning.  From this perspective, we conduct a thorough examination of recent advancements in world models, analyzing them through the following lenses, as depicted in Figure~\ref{fig:intro}.

\begin{itemize}
    \item \textbf{Implicit representation of the external world} (Section~\ref{sec::inner}): This research category constructs a model of environmental change to enable more informed decision-making, ultimately aiming to predict the evolution of future states. It fosters an implicit comprehension by transforming external realities into a model that represents these elements as latent variables. Furthermore, with the advent of large language models (LLMs), efforts previously concentrated on traditional decision-making tasks have been significantly enhanced by the detailed descriptive power of these models regarding world knowledge. We further focus on the integration of world knowledge into existing models.

    \item \textbf{Future predictions of the external world} (Section~\ref{sec::future}): We initially explore generative models that simulate the external world, primarily using visual video data. These works emphasize the realness of generated videos that mirror future states of the physical world. As recent advancements shift focus toward developing a truly interactive physical world, we further investigate the transition from visual to spatial representations and from video to embodiment. This includes comprehensive coverage of studies related to the generation of embodied environments that mirror the external world.

    \item \textbf{Applications of world models} (Section~\ref{sec:application}): World models have demonstrated wide-ranging applications across diverse domains, spanning game intelligence, embodied agents, urban systems, and societal modeling. These domains—represented respectively by generative games, robotics, autonomous driving, and social simulacra—illustrate how world models bridge perception, reasoning, and imagination across both virtual and physical environments. We explore how the integration of world models in these domains advances theoretical understanding and practical innovation alike, underscoring their transformative potential in shaping intelligent systems.

\end{itemize}

\section{Implicit Representation of the External World}
\label{sec::inner}

This section examines how world models enable informed decision-making by representing the environment as latent variables. Section~\ref{sec::decisionmaking} focuses on the world models in model-based RL (MBRL), while Section~\ref{world_knowledge} explores the integration of world knowledge into advanced AI models, especially LLMs, enhancing real-world task performance.

\subsection{World Model in Decision-Making} \label{sec::decisionmaking}
\begin{figure}[h]
    \centering
    \includegraphics[width=0.99\linewidth]{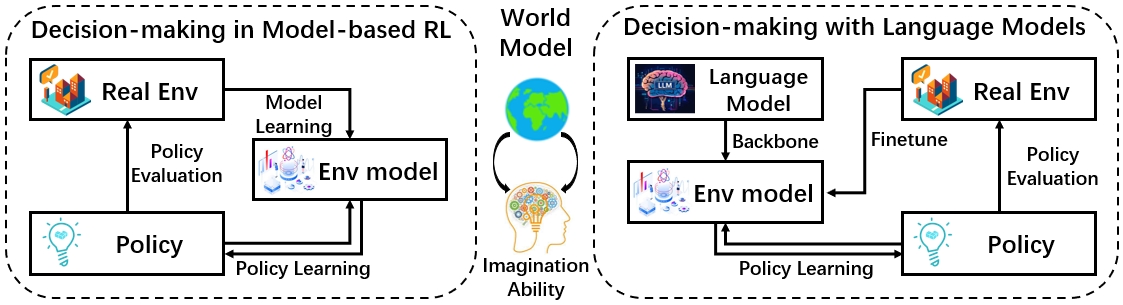}
    \caption{Two schemes of utilizing world model in decision-making.}
    \label{fig:decisionmaking}
\end{figure}

In decision-making tasks, understanding the environment is the major task in setting a foundation for optimized policy generation. As such, the world model in decision-making should include a comprehensive understanding of the environment. It enables us to take hypothetical actions without affecting the real environment, facilitating a low trial-and-error cost. In literature, research on how to learn and utilize the world model was initially proposed in the field of model-based RL. Furthermore, recent progress on LLM and MLLM also provide comprehensive backbones for world model construction. With language serving as a more general representation, language-based world models can be adapted to more generalized tasks. The two schemes of leveraging world models in decision-making tasks are shown in Figure~\ref{fig:decisionmaking}.

\subsubsection{World model in model-based RL}
In decision-making, the concept of the world model largely refers to the environment model in MBRL. A decision-making problem is typically formulated as a Markov Decision Process (MDP), denoted with a tuple $(S, A, M, R, \gamma)$, where $S, A, \gamma$ denotes the state space, action space and the discount factor each. The world model here consists of $M$, the state transition dynamics and $R$, the reward function. Since the reward function is defined in most cases, the key task of MBRL is to learn and utilize the transition dynamics, which can further support policy optimization.

\textbf{World Model Learning.} To learn an accurate world model, the most straightforward approach is to leverage the mean squared prediction error on each one-step transitions~\cite{kurutach2018model, luo2018algorithmic, janner2019trust, rajeswaran2020game, janner2021offline}, 

\begin{equation}
    \min_\theta \mathbb{E}_{s' \sim M^{*}(\cdot|s,a)}[||s' - M_\theta(s,a)||^2_2],
\end{equation}
where $M^*$ is the real transition dynamics used to collect trajectory data and $M_\theta$ is the parameterized transition to learn. Apart from directly utilizing the deterministic transition model, Chua et al.\cite{chua2018deep} further model the aleatoric uncertainty with the probabilistic transition model. The objective is to minimize the KL divergence between the transition models,
\begin{equation}
    \min_\theta \mathbb{E}_{s' \sim M{*}(\cdot|s,a)}[log(\frac{M^*(s'|s,a)}{M_\theta(s'|s,a)})].
\end{equation}
In both settings, the phase of the world model learning task can be transformed into a supervised learning task. The learning labels are the trajectories derived from real interaction environments, also called the simulation data~\cite{luo2024survey}.

For high-dimensional environments, representation learning is essential for effective world-model training in MBRL. Early work by Ha and Schmidhuber\cite{ha2018recurrent} reconstructs images through an autoencoder–latent-state pipeline, whereas Hafner et al. \cite{hafner2019dream, hafner2020mastering} couple a visual encoder with latent dynamics to master pixel-based control tasks. Their latest iteration, DreamerV3\cite{hafner2025mastering}, adds robust normalization and balancing techniques, solving over 150 tasks—including diamond collection in Minecraft—without human data or domain-specific tuning. Memory-centric extensions such as Recall-to-Imaging by Samsami et al.\cite{samsami2024mastering} further enhance long-horizon reasoning. A complementary trend is unified model learning via next-token prediction with transformer architectures, as shown by Janner et al. \cite{janner2021offline} and expanded by Schubert et al. \cite{schubert2023generalist}. Further, Georgiev et al. \cite{georgievpwm} train a large off-policy multi-task world model whose smooth latent dynamics enable efficient per-task policy learning with first-order gradients, achieving strong scalability and performance without online planning.
Recent work by Jonathan Richens et al.\cite{richens2025general} further reinforces the necessity of world models, showing that any agent capable of generalizing to multi-step goal-directed tasks must have learned a predictive model of its environment, with the world model emerging from the agent's policy. This insight aligns with the ongoing trend of incorporating predictive modeling into reinforcement learning to handle more complex and goal-oriented tasks.

\textbf{Policy Generation with World Model.} With an ideally optimized world model, one most straightforward way to generate a corresponding policy is model predictive control (MPC)\cite{kouvaritakis2016model}. MPC plans an optimized sequence of actions given the model as follows:
\begin{equation}
    \max_{a_{t:t+\tau}}\mathbb{E}_{s_{t'+1}\sim p(s_{t'+1}|s_{t'}, a_{t'})}[\sum^{t+\tau}_{t'=t}r(s_{t'}, a_{t'})],
\end{equation}
where $\tau$ denotes the planning horizon. Nagabandi et al.\cite{nagabandi2018neural} adopt a simple Monte Carlo method to sample action sequences. Rather than sampling actions uniformly, Chua et al.\cite{chua2018deep} propose a new probabilistic algorithm that ensembles with trajectory sampling. Further literature also improves the optimization efficiency by leveraging the world model usage~\cite{hafner2019dream, yu2016derivative, hu2017sequential,wang2019exploring}. Hansen et al.\cite{hansentd} introduced an improved model-based RL algorithm called TD-MPC2 that integrates trajectory optimization within the latent space of a learned implicit world model. It achieves strong performance across diverse continuous control tasks and demonstrates scalability by training large agents with hundreds of millions of parameters across multiple domains.

Another popular approach to generating world model policies is the Monte Carlo Tree Search (MCTS). By maintaining a search tree where each node refers to a state evaluated by a predefined value function, actions will be chosen such that the agent can be processed to a state with a higher value. AlphaGo and AlphaGo Zero are two significant applications using MCTS in discrete action space~\cite{silver2016mastering, silver2017mastering}. Moerland et al.~\cite{moerland2018a0c} extended MCTS to solve decision problems in continuous action space. Oh et al.~\cite{oh2017value} proposed a value prediction network that applies MCTS to the learned model to search for actions based on value and reward predictions. 

\subsubsection{World model with language backbone}
The rapid growth of language models, especially LLM and MLLM, benefits development in many related applications. With language serving as a universal representation backbone, language-based world models have shown their potential in many decision-making tasks.

\textbf{Direct Action Generation via LLM World Models.}
LLM is capable of directly generating actions in decision-making tasks based on corresponding constructed world models. For example, in the navigation scenarios, Yang et al.~\cite{yang2023probabilistic} transfer pre-trained text-to-video models to domain-specific tasks for robot control, successfully annotating robot manipulation with text instructions as LLM outputs.  Zhou et al.~\cite{zhou2024robodreamer} further learn a compositional world model by factorizing the video generation process. Such a method enables a strong few-shot transfer ability to unseen tasks. 

Besides training or fine-tuning specialized language-based world models, LLMs and MLLMs can be directly deployed to understand the world environment in decision-making tasks. For example, Long et al.~\cite{long2024discuss} propose a multi-expert scheme to handle visual language navigation tasks. They construct a standardized discussion process where eight LLM-based experts participate to generate the final movement decision. An abstract world model is constructed from the discussion and further imagination (of future states) of the experts to support action generation. Zhao et al.~\cite{zhao2024over} further combine LLMs and open-vocabulary detection to construct the relationship between multi-modal signals and key information in navigation. They propose an omni-graph to capture the structure of the local space as the world model for the navigation task. Meanwhile, Yang et al.~\cite{yang2024rila} utilize an LLM-based imaginative assistant to infer the global semantic graph as the world model based on the environment perception, and another reflective planner to directly generate actions.

Recent works have continued to enhance this paradigm by addressing specific challenges, such as those found in web navigation, as exemplified by the research on web agents~\cite{chae2024web, qiao2024agent}. Chae et al.~\cite{chae2024web} developed a World-model-augmented (WMA) agent that improves web navigation by predicting action outcomes via a novel transition-focused observation abstraction, addressing complex issues like irreversible actions such as purchasing non-refundable flight tickets. Similarly, Qiao et al.~\cite{qiao2024agent} proposed a parametric World Knowledge Model (WKM) that provides agents with both prior global knowledge and dynamic local knowledge, effectively mitigating common problems like blind trial-and-error and hallucinatory actions.

\textbf{Modular Usage of LLM World Models.}
Although taking LLM outputs as actions directly is straightforward in application and deployment, the decision quality in such a scheme heavily relies on the reasoning ability of the LLM itself. 
Although this year has witnessed the large potential of LLM reasoning capability~\cite{xu2025towards}, it can be further improved by integrating LLM-based world models as modules with external model-based verifiers or other effective planning algorithms~\cite{kambhampati2024position}. 

Guan et al.\cite{guan2023leveraging} extract explicit world models by prompting GPT-4 to generate and iteratively refine PDDL domain descriptions, then pair these models with off-the-shelf planners, yielding good planning performance with much less human intervention.
Xiang et al.\cite{xiang2024language} deploys an embodied agent in a world model, the simulator of VirtualHome~\cite{puig2018virtualhome}, where the corresponding embodied knowledge is injected into LLMs. To better plan and complete specific goals, they propose a goal-conditioned planning schema where Monte Carlo Tree Search (MCTS) is utilized to search for the true embodied task goal. Lin et al.\cite{lin2024learningmodelworldlanguage} introduce an agent, Dynalang, which learns a multimodal world model to predict future text and image representations, and which learns to act from imagined model rollouts. The policy learning stage utilizes an actor-critic algorithm purely based on the previously generated multimodal representations. Liu et al.~\cite{liu2024reasonfutureactnow} further cast reasoning in LLMs as learning and planning in Bayesian adaptive Markov decision processes (MDPs). LLMs, like the world model, perform in an in-context manner within the actor-critic updates of MDPs. The proposed RAFA framework shows significantly increased performance in multiple complex reasoning tasks and environments, such as ALFWorld~\cite{shridhar2020alfworld}. 

This modular approach has also been successfully applied to specific domains, such as web navigation. Gu et al.~\cite{gu2024your} proposed WebDreamer, a model-based planning framework that uses a specialized LLM as a world model to simulate actions, achieving competitive performance with significantly higher efficiency than tree-search methods on the web. In a different approach, Tang et al.~\cite{tang2024worldcoder} introduced WorldCoder, a model-based agent that builds and refines its world model by writing and editing a Python program, demonstrating greater sample and compute efficiency than existing methods.

\subsection{World Knowledge Learned by Models} \label{world_knowledge}

\begin{table}
\centering
\caption{Overview of recent works in world knowledge learned by models.}\label{tbl:world}
\resizebox{\textwidth}{!}{
\begin{tabular}{lllll} 
\toprule
\textbf{Category} & \textbf{Methods/Model} & \textbf{Year\&Venue} & \textbf{Modality} & \textbf{Content} \\ 
\midrule
\multirow{6}{*}{\textbf{Common Sense \& General Knowledge}} 
 & KoLA~\cite{yu2023kola} & 2024 ICLR & Language & Benchmark \\
 & EWOK~\cite{ivanova2024elements} & 2024 arxiv & Language & Benchmark \\
 & Geometry of Concepts~\cite{li2024geometryconceptssparseautoencoder} & 2024 arxiv & Language & Analysis \\
 & BLEnD~\cite{myung2024blend} & 2024 NeurIPS & Language & Benchmark \\
 & PIGEON~\cite{lan2025open} & 2025 ACL Findings & Language & Prediction \\
 & LocalGPT~\cite{lan2025benchmarking} & 2025 KDD & Language & Benchmark \\
 \midrule
\multirow{10}{*}{\textbf{Knowledge of Global Physcial World}} & Space\&Time~\cite{gurnee2023language} & 2024 ICLR & Language & Analysis \\
 & GeoLLM~\cite{manvi2023geollm} & 2024 ICLR & Language & Understanding \\
 & GeoLLM-Bias~\cite{manvi2024large} & 2024 ICML & Language & Understanding \\
 & GPT4GEO~\cite{roberts2023gpt4geo} & 2023 NeurIPS(FMDM) & Language & Benchmark \\
 & CityGPT~\cite{feng2024citygpt} & 2025 KDD & Language & Understanding \\
 & CityBench~\cite{feng2024citybench} & 2025 KDD & Language\&Vision & Benchmark \\
 & UrbanLLaVA~\cite{feng2025urbanllava} & 2025 ICCV & Language\&Vision & Understanding \\
 & GPS-To-Image~\cite{feng2025gps} & 2025 CVPR & Vision & Generation \\
 & Ai's Blind Spots~\cite{beneduce2025ai} & 2025 arxiv & vision & Generation \\
 & AgentMove~\cite{feng2025agentmove} & 2025 NAACL & Language & Prediction \\
 & GLOBE~\cite{li2025recognition} & 2025 arxiv & Language\&Vision & Understanding \\
\midrule
\multirow{6}{*}{\textbf{Knowledge of Local Physical World}} 
 & Predictive~\cite{gornet2024automated} & 2024 NMI & Vision & Learning \\
 & Emergent~\cite{jinemergent} & 2024 ICML & Language & Learning \\
 & E2WM~\cite{xiang2024language} & 2023 NeurIPS & Language & Learning \\
 & Dynalang~\cite{lin2024learningmodelworldlanguage} & 2024 ICML & Language\&Vision & Learning \\ 
 & WM-ABench~\cite{gao2025vision} & 2025 ACL Findings & Vision & Benchmark \\
 & Spatial457~\cite{spatiallm} & 2025 CVPR & Vision & Benchmark \\
 & Thinking in Space~\cite{yang2025thinking} & 2025 CVPR & Vision & Benchmark \\
\midrule
\multirow{9}{*}{\textbf{Knowledge of Human Society}} & Testing ToM~\cite{strachan2024testing} & 2024 NHB & Language & Benchmark \\
 & High-order ToM~\cite{street2024llms} & 2024 arxiv & Language & Benchmark \\
 & COKE~\cite{wu-etal-2024-coke} & 2024 ACL & Language & Learning \\
 & MuMA-ToM~\cite{shi2024muma} & 2024 ACL & Language\&Vision & Benchmark \\
 & SimToM~\cite{wilf2023thinktwiceperspectivetakingimproves} & 2024 ACL & Language & Learning \\
 & EAI~\cite{mozikov2024eai} & 2024 NeurIPS & Language & Benchmark \\
 & LLM-ToM~\cite{kosinski2024evaluating} & 2024 PNAS & Lanauge & Benchmark \\
  & SafeWorld~\cite{yin2024safeworld} & 2024 NeurIPS & Lanuage & Benchmark \\
 & 100 languages~\cite{vayani2025all} &2025 CVPR& Language & Benchmark \\
\bottomrule
\end{tabular}}
\end{table}

\begin{figure}[h]
    \centering
    \includegraphics[width=0.9\linewidth]{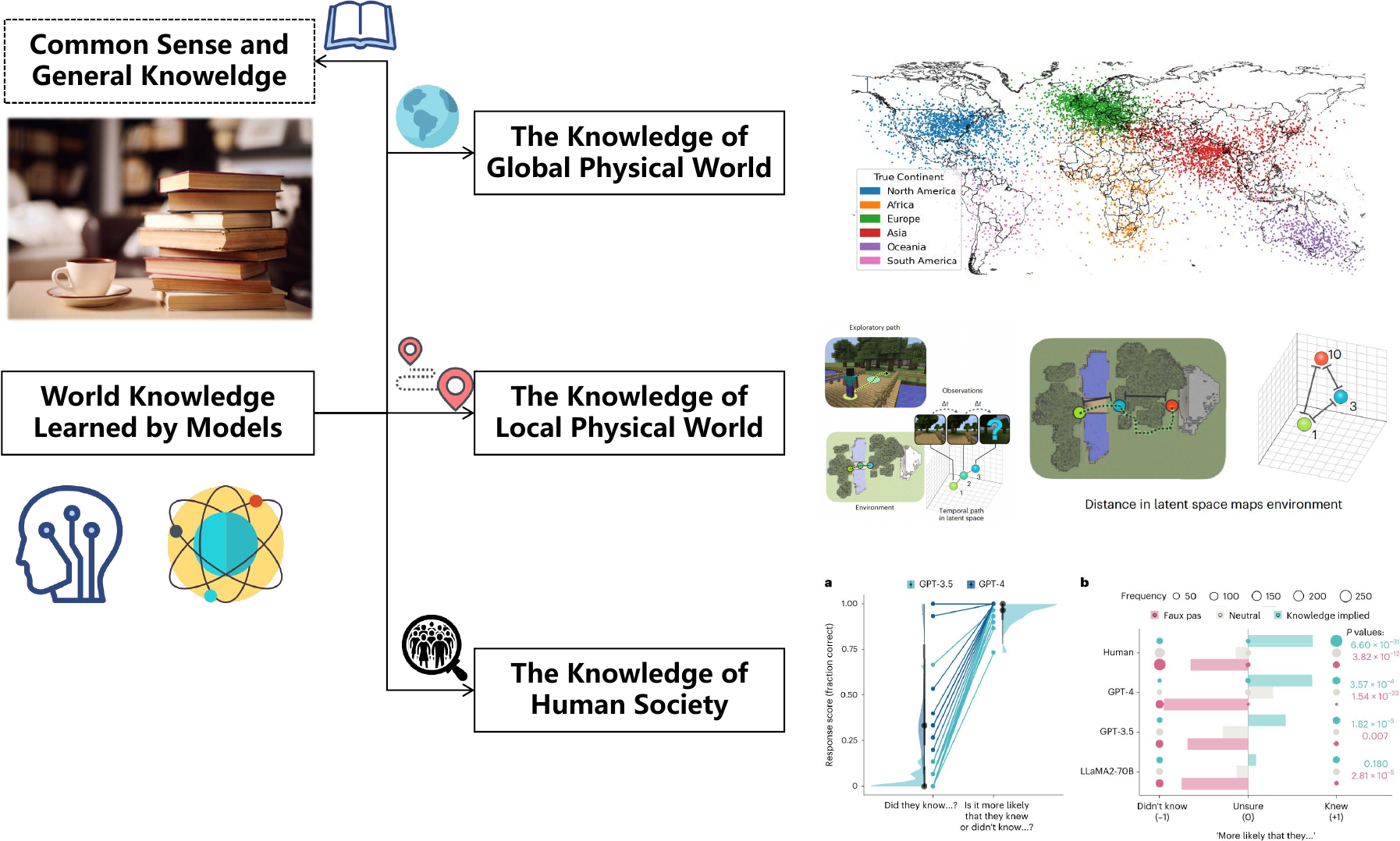}
    \caption{World knowledge in large language models for world model.}
    \label{fig:worldknowledge}
\end{figure}

After pretraining on large-scale web text and books~\cite{touvron2023llama, chatgpt}, large language models attain extensive knowledge about the real world and common sense relevant to daily life. This embedded knowledge is considered crucial for their remarkable ability to generalize and perform effectively in real-world tasks. For instance, researchers leverage the common sense of large language models for task planning~\cite{zhao2024large}, robot control~\cite{huang2022inner}, and image understanding~\cite{liu2024visual}. Furthermore, Li et al.\cite{li2024geometryconceptssparseautoencoder} discover brain-like structures of world knowledge embedded in the high-dimensional vectors that represent the universe of concepts in large language models. Also, Li et al.\cite{li2024vision} demonstrate that language models partially converge towards representations isomorphic to those of vision models. Building upon this extensive knowledge of human daily life~\cite{myung2024blend}, LLMs have been successfully applied in real-world scenarios. For example, by leveraging this prior knowledge to provide semantic information for everyday human activities, they have proven effective in domains such as local life services~\cite{lan2025benchmarking, lan2025open}.

Unlike common sense and general knowledge, we focus on world knowledge within large language models from the perspective of a world model. As shown in Figure~\ref{fig:worldknowledge}, based on objects and spatial scope, the world knowledge in the large language models can be categorized into three parts: 1) knowledge of the global physical world; 2) knowledge of the local physical world; and 3) knowledge of human society. We summarize recent works in Table~\ref{tbl:world}.

\subsubsection{Knowledge of the Global Physical World}
We first introduce research focused on analyzing and understanding the knowledge of the global physical world. 
Gurnee et al.~\cite{gurnee2023language} present the first evidence that large language models genuinely acquire spatial and temporal knowledge of the world, rather than merely collecting superficial statistics. They identify distinct "spatial neurons" and "temporal neurons" in LLama2~\cite{touvron2023llama}, suggesting that the model learns linear representations of space and time across multiple scales. Distinct from previous observations focused on embedding space, Manvi et al.\cite{manvi2023geollm, manvi2024large} develop effective prompts about textual address to extract intuitive real-world knowledge about the geospatial space and successfully improve the performance of the model in various downstream geospatial prediction tasks. 

While large language models do acquire some implicit knowledge of the real world~\cite{gurnee2023language, li2024geometryconceptssparseautoencoder}, the quality of this knowledge remains questionable~\cite{roberts2023gpt4geo, feng2024citygpt}. For example, Feng et al.~\cite{feng2024citygpt, feng2025urbanllava} find that the urban knowledge embedded in large language models is often coarse and inaccurate. To address this, they propose an effective framework to improve the acquisition of urban knowledge of specific cities in large language models. Building on the global geospatial knowledge embedded in LLMs, researchers are now applying this prior world knowledge to overcome the generalization challenges faced by previous methods, for instance, AgentMove~\cite{feng2025agentmove} for global mobility prediction, GPS-to-Image~\cite{feng2025gps, beneduce2025ai} for generating geographically-aware and stylistically-controllable images, and GLOBE for knowledge based image geo-localization~\cite{li2025recognition}.

From a long-term perspective, we can see that although large language models have demonstrated the ability to capture certain aspects of real-world knowledge~\cite{gurnee2023language, li2024geometryconceptssparseautoencoder, roberts2023gpt4geo}, it is clear that further efforts are needed to enhance this knowledge to enable broader and more reliable real-world applications~\cite{feng2025survey}. 

\subsubsection{Knowledge of the Local Physical World}
Unlike the knowledge of the global physical world, the local physical world represents the primary environment for human daily life and most real-world tasks. Therefore, understanding and modeling the local physical world is a more critical topic for building a comprehensive world model. We first introduce the concept of the cognitive map~\cite{tolman1948cognitive}, which refers to the mental representation that humans form to navigate and understand their environment, including spatial relationships and landmarks. Although initially developed to explain human learning processes, researchers have discovered similar structures in large language models~\cite{li2024geometryconceptssparseautoencoder} and have leveraged these insights to enhance the efficiency and performance of artificial models in learning and understanding the physical world.

Recent studies explore actively encouraging models to learn abstract knowledge through cognitive map-like processes across various environments. For example, Cornet et al.~\cite{gornet2024automated} show that in a simplified Minecraft world, visual predictive coding lets an agent build a spatial cognitive map purely from pixels. Once trained, the latent map encodes its metric distance to any target, enabling accurate rollout of future observations. Lin et al.~\cite{lin2024learningmodelworldlanguage} investigate teaching models to understand the game environments through a world model learning procedure, specifically by predicting the subsequent frame of the environment. In this way, the model can generate better actions in dynamic environments. Moreover, Jin et al.~\cite{jinemergent} find that language models can learn the emergent representations of program semantics by predicting the next token. Recently, researchers~\cite{gao2025vision,spatiallm,yang2025thinking} have extended these studies to more realistic settings, revealing a substantial gap in the ability of large language model-based methods to construct precise models, even for simple local environments.

\subsubsection{Knowledge of the Human Society}
Beyond the physical world, understanding human society is another crucial aspect of world models. David Premack and Guy Woodruff proposed the Theory of Mind~\cite{premack1978does}, which was later developed to explain how individuals infer the mental states of others around them. Recent works have extensively explored how large language models develop and demonstrate this social world model~\cite{sap2022neural,strachan2024testing,kosinski2024evaluating}. Sap et al.~\cite{sap2022neural} conduct an investigation focusing on evaluating the performance of large language models across various Theory of Mind tasks to determine whether their human-like behaviors reflect genuine comprehension of social rules and implicit knowledge. Strachan et al.~\cite{strachan2024testing} conduct a comparative analysis between human and LLM performance on diverse Theory of Mind abilities, such as understanding false beliefs and recognizing irony. While their findings demonstrate the potential of GPT-4 in these tasks, they also identify its limitations, particularly in detecting faux pas. 

Beyond inferring individual mental states, researchers are also investigating how LLMs model the broader, underlying rules of human society. For instance, Mozikov et al.~\cite{mozikov2024eai} investigated how emotional factors influence ethical judgment and decision-making in LLMs, underscoring the need for robust mechanisms to ensure consistent ethical standards. Other works have explored the capacity of these models to navigate the complexities of a globalized world. Yin et al.~\cite{yin2024safeworld}, for example, evaluated LLMs on their ability to generate responses that are not only helpful but also culturally sensitive and legally compliant across diverse global contexts. Similarly, Vayani~\cite{vayani2025all} conducted a large-scale evaluation of LLMs across a hundred culturally diverse languages, highlighting the importance of linguistic diversity in the development of truly global social models.

While these studies demonstrate the great potential of LLMs in modeling the social world, they also reveal significant limitations when these models must handle complex social situations. To address these shortcomings and enhance LLMs' Theory of Mind abilities for complex, real-world applications, researchers have proposed several innovative methods. For instance, Wu et al.~\cite{wu-etal-2024-coke} introduced COKE, a framework that constructs a knowledge graph to help LLMs explicitly apply Theory of Mind through cognitive chains. Additionally, Alex et al.~\cite{wilf2023thinktwiceperspectivetakingimproves} developed SimToM, a two-stage prompting framework designed to improve the performance of large language models on Theory of Mind tasks.

\section{Future Prediction of the Physical World} \label{sec::future}

\subsection{World Model as Video Generation}  
  
The integration of video generation into world models marks a significant leap forward in the field of environment modeling~\cite{sora2024}. Traditional world models primarily focused on predicting discrete or static future states~\cite{ha2018world,lecun2022path}. However, by generating video-like simulations that capture continuous spatial and temporal dynamics, world models~\cite{sora2024,yang2024worldgpt} have evolved to address more complex, dynamic environments. This breakthrough in video generation has pushed the capabilities of world models to a new level.

\subsubsection{Towards Video World Models}

A video world model is a computational framework designed to simulate and predict the future state of the world by processing past observations and potential actions within a visual context~\cite{sora2024}. This concept builds on the broader idea of world models, which strive to capture the dynamics of an environment and enable machines to predict how the world will evolve over time. In the case of a video world model, the focus is on generating sequences of visual frames that represent these evolving states.

Sora~\cite{sora2024} is a large-scale video generation model designed for high-quality, temporally consistent video sequences up to one minute long, based on various multimodal inputs. It leverages neural network architectures to produce visually coherent simulations that often align with real-world physical principles, like light reflection or melting. These capabilities suggest Sora's potential as a world simulator, predicting future states based on initial conditions and parameters. However, despite its impressive video generation, Sora has significant limitations in fully understanding and simulating the external world. A key limitation is its causal reasoning ability~\cite{zhu2024sora,cho2024sora}, restricting it to passively generating sequences without actively predicting how actions might alter events. Furthermore, Sora struggles to consistently reproduce correct physical laws~\cite{kang2024far}, failing to accurately simulate complex physics like object behavior under forces, fluid dynamics, or light interactions.

Following the success of Sora in generating high-quality videos, the past two years have witnessed the emergence of several large-scale video generative base models, such as OpenSora~\cite{zheng2024open}, CogVideoX~\cite{yang2024cogvideox}, and Wan~\cite{wan2025wan}. These models, through the pre-training of more efficient VAEs and extensive pre-training on large-scale video datasets, exhibit strong visual generation capabilities and serve as foundational components for world models. Further advancing this field, Cosmos~\cite{agarwal2025cosmos} introduces a dedicated video generation base model for physical world simulation, achieving new breakthroughs in physical law adherence and understanding by pre-training on massive real-world physics videos and exploring both diffusion and auto-regressive architectures. Concurrently, Genie 2~\cite{parkerholder2024genie2} and Genie 3~\cite{genie3} focus on video generation in gaming scenarios, with Genie 2 specifically designing an auto-regressive diffusion architecture for interactive video generation that supports following external action instructions. Beyond these specific models, continuous progress is being made in key technical challenges including long-duration generation~\cite{yin2023nuwa,liu2024world,hu2023gaia,henschel2025streamingt2v}, interactive generation~\cite{zhen20243d,xiang2024pandora,yang2023learning,yang2024video,wu2024ivideogpt,zhang2025physdreamer,jain2024peekaboo,xiang2024pandora,team2025aether,mao2025yume,bar2025navigation}, and physical law adherence~\cite{yang2024worldgpt, cai2023diffdreamer,ren2024consisti2v, shang2025roboscape}. 
Researchers are increasingly shifting focus from basic, user-uncontrolled video generation towards interactive simulations that replicate real-world decision spaces to facilitate decision-making. Moreover, the concept of world models has expanded beyond pure imagination, finding application in diverse scenario-specific simulations~\cite{liu2024world, wang2024worlddreamer, bruce2024genie, mendonca2023structured, hu2023gaia, wang2023drivedreamer, bogdoll2023muvo, min2023uniworld}, encompassing natural environments, games, autonomous driving, and robotics.

\begin{table}[t]
\footnotesize
\centering
\caption{Overview of recent models in video generation across various categories, which summarizes key models in long-term video generation, multi-modal learning, interactive video generation, temporal consistency, and diverse environment modeling.}\label{tbl:video}
\resizebox{\textwidth}{!}{
\begin{tabular}{>{\raggedright\arraybackslash}p{2cm}|c|>{\raggedright\arraybackslash}p{6cm}|>{\raggedright\arraybackslash}p{2cm}}
\toprule
\textbf{Category} & \textbf{Model} & \textbf{Description} & \textbf{Technique} \\
\midrule
\multirow{3}{2cm}{Long-term} &     NUWA-XL~\cite{yin2023nuwa}       &   ``Coarse-to-fine'' Diffusion over Diffusion architecture for long video generation.    &     Diffusion  \\
&     LWM~\cite{liu2024world}       &     Training large transformers on long video and language sequences.      &  Transformer    \\
&     GAIA-1~\cite{hu2023gaia}       &   Generative world model predicting driving scenarios for autonomous driving.     &    Transformer, Diffusion    \\
&     StreamingT2V~\cite{henschel2025streamingt2v}       &   An autoregressive text-to-video model equipped with long/short-term memory blocks.     &   Diffusion \\
\midrule
\multirow{3}{2cm}{Multimodal}  &    3D-VLA~\cite{zhen20243d}       &  Integrates 3D perception, reasoning, and action in a world model for embodied AI.   &  Diffusion    \\
& Pandora~\cite{xiang2024pandora}     & World-state simulation and real-time control with free-text actions.   &   LLM     \\
&   Genie~\cite{bruce2024genie}    &  Generative model from text, images, and sketches.  &   Transformer    \\
\midrule
\multirow{5}{2cm}{Interactive}  &  UniSim~\cite{yang2023learning}   &  Simulates real-world interactions for vision-language and RL training.    &   Diffusion,   RL   \\
&  VideoDecision\cite{yang2024video}   &  Extends video models to real-world tasks like planning and RL.   &    Transformer, Diffusion               \\
&  iVideoGPT~\cite{wu2024ivideogpt}   & Combines visual, action, and reward signals for interactive world modeling.   &    Transformer               \\
&  PEEKABOO~\cite{jain2024peekaboo}   &Enhances interactivity with spatiotemporal control without extra training.  &     Diffusion Transformer              \\
&  Aether~\cite{team2025aether}   & Utilizes camera trajectories as geometry-aware actions, enabling accurate action-conditioned prediction and visual planning.  &     Diffusion       \\
&  Yume~\cite{mao2025yume}   & Enhances streaming interactive world generation following continuous keyboard inputs. &     Diffusion            \\
&  NWM~\cite{bar2025navigation}   & Achieves controllable video generation based on past observations and navigation actions.  &     Diffusion              \\
\midrule
\multirow{3}{2cm}{Consistency}  &  WorldGPT~\cite{yang2024worldgpt}     &   Improves temporal consistency and action smoothness with multimodal learning and refined key frame generation. &      Diffusion             \\
&  DiffDreamer~\cite{cai2023diffdreamer}    &    Long-range scene extrapolation with improved consistency.   &   Diffusion   \\
&  ConsistI2V~\cite{ren2024consisti2v}   & Enhances visual consistency in image-to-video generation. &   Diffusion  \\
&  WorldMem~\cite{xiao2025worldmem}   & Enhances long-term consistent world simulation with an integrated memory mechanism. &   Diffusion  \\
\midrule
\multirow{4}{2cm}{Diverse environments}    &   WorldDreamer~\cite{wang2024worlddreamer}    &  World model capturing dynamic elements across diverse scenarios.  &   Transformer \\
&   Genie~\cite{bruce2024genie}    &  Unsupervised generative model for action-controllable virtual environments.  &   Transformer    \\
&   MUVO~\cite{bogdoll2023muvo}   &  Multimodal world model using camera and lidar data.  &    Transformer  \\
&   UniWorld~\cite{min2023uniworld}     &  3D detection and motion prediction in autonomous driving.  &   Transformer  \\
\bottomrule
\end{tabular}}
\end{table}

\subsubsection{Capabilities of Video World Models}

Despite the ongoing debate about whether models like Sora can be considered full-fledged world models, there is no doubt that video world models hold tremendous potential for advancing environment simulation and prediction~\cite{zhu2024sora,cho2024sora,kang2024far}. These models can offer a powerful approach to understanding and interacting with complex environments by generating realistic, dynamic video sequences. To achieve this level of sophistication, this section outlines the key capabilities that video world models must possess to set them apart from traditional video generation models.

\textbf{Long-Term Predictive Ability.} A robust video world model should be capable of making long-term predictions that adhere to the dynamic rules of the environment over an extended period. This capability allows the model to simulate how a scenario evolves, ensuring that the generated video sequences remain consistent with the temporal progression of the real world. Although Sora has achieved the generation of minute-long video sequences with high-quality temporal coherence, it is still far from being able to simulate complex, long-term dynamics found in real-world environments. Recent efforts have explored extending video lengths to capture longer-term dependencies and improve temporal consistency~\cite{yin2023nuwa,liu2024world,hu2023gaia}.

\textbf{Multi-Modal Integration.}  In addition to language-guided video generation, video world models are increasingly integrating other modalities, such as images and actions, to enhance realism and interactivity~\cite{zhen20243d,xiang2024pandora}. The integration of multiple modalities allows for richer simulations that better capture the complexity of real-world environments, improving both the accuracy and diversity of generated scenarios.

\textbf{Interactivity.} Another critical capability of video world models is their potential for controllability and interactivity. An ideal model should not only generate realistic simulations but also allow for interaction with the environment. This interactivity involves simulating the consequences of different actions and providing feedback, enabling the model to be used in applications requiring dynamic decision-making. Recent work is focusing on enhancing control over the simulations, allowing for more user-guided exploration of scenarios~\cite{yang2024video,wu2024ivideogpt}.

\textbf{Diverse Environments.} Finally, video world models are being adapted to a variety of scenario-specific simulations, including natural environments, autonomous driving, and gaming. These models are evolving beyond basic video generation to replicate real-world dynamics and support a wide range of applications~\cite{liu2024world, wang2024worlddreamer, bruce2024genie}.

\subsection{World Model as Embodied Environment}
The development of world models for embodied environments is crucial for simulating and predicting how agents interact with and adapt to the external world. Initially, generative models focused on simulating visual aspects of the world, using video data to capture dynamic changes in the environment. More recently, the focus has shifted towards creating fully interactive and embodied simulations. These models not only represent the visual elements of the world but also incorporate spatial and physical interactions that more accurately reflect real-world dynamics. By integrating spatial representations and transitioning from video-based simulations to immersive, embodied environments, world models can now provide a more comprehensive platform for developing agents capable of interacting with complex real-world environments.

\begin{figure}[!t]
    \centering
    \includegraphics[width=0.9\linewidth]{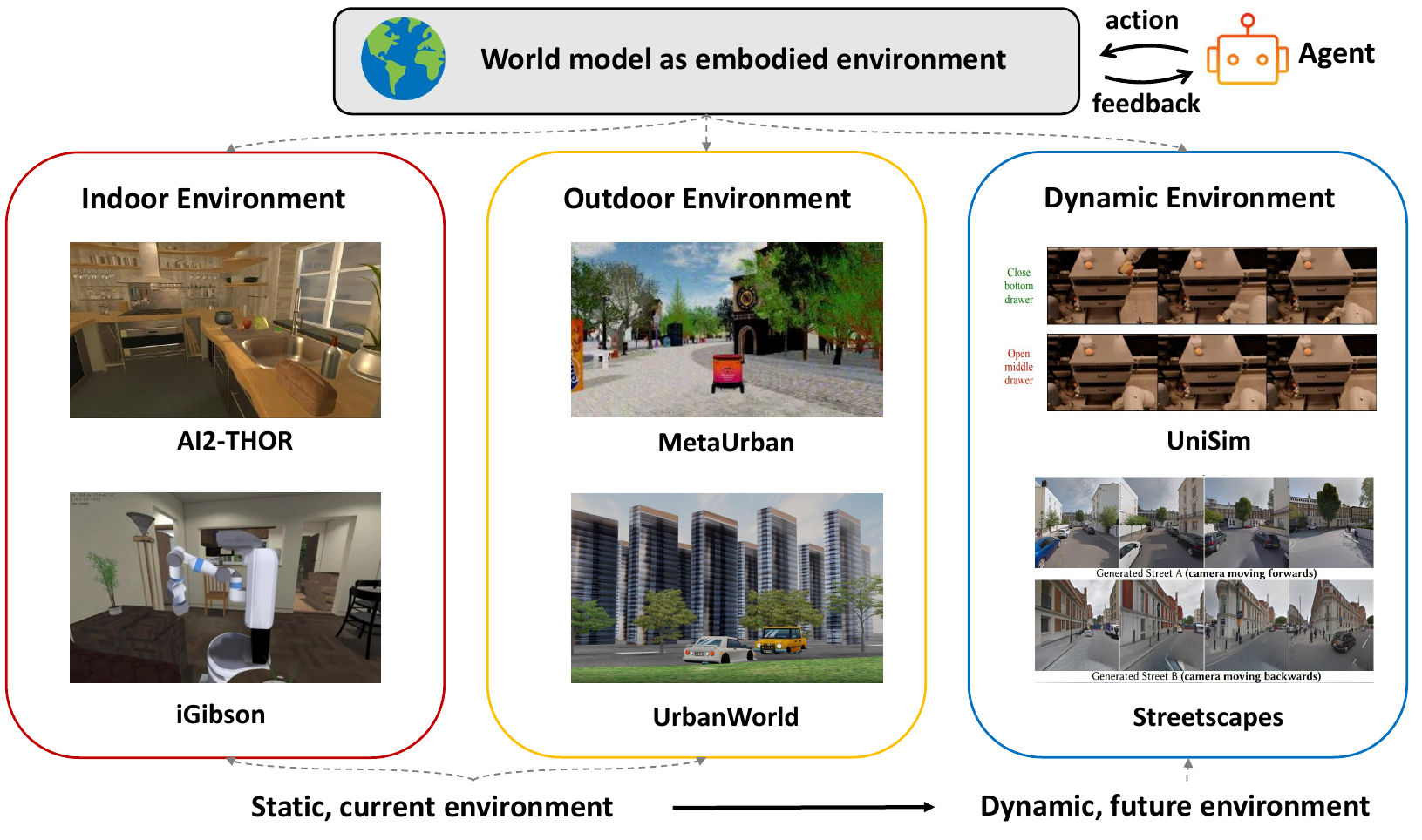}
    \caption{Classification of world models as interactive embodied environments, including indoor, outdoor and dynamic environments. The modeling of the outside world is evolving from constructing static, current environments to predicting dynamic, future environments.}
    \label{fig:embodied env}
\end{figure}

World models as embodied environments can be divided into three categories: indoor, outdoor, and dynamic environments, as shown in Figure~\ref{fig:embodied env}, and the relevant works are summarized in Table~\ref{embody-env}.
It can be summarized that most current works focus on developing static, existing indoor and outdoor embodied environments.
An emerging trend is to predict the dynamic, future world through generative models producing first-person, dynamic video-based simulation environments. 
Such environments can offer flexible and realistic feedback for training embodied agents, enabling them to interact with ever-changing environments and improve their generalization ability.
\begin{table}[t]
\footnotesize
\centering
\caption{\label{comp}Comparison of existing works on world models as embodied environments, including indoor, outdoor, and dynamic environments. In the `Modality' column, `V' refers to vision, `L' refers to lidar, `T' refers to text, and `A' refers to audio. In the `Num of Scenes' column, `-' means no reported data, and  `Arbitrary' means the method can support generating any number of scenes.}
\resizebox{\linewidth}{!}{
\begin{tabular}{c|c|c|c|c|c|c|c}
\hline
\textbf{Type} & \textbf{Name} & \textbf{Environment} & \textbf{Year} & \textbf{Num of Scenes} & \textbf{Modality} & \textbf{Physics} & \textbf{3D Assets} \\
\hline
Indoor & AI2-THOR~\cite{kolve2017ai2}   & Home    & 2017 & 120    & V   & \checkmark & \checkmark \\
Indoor & Matterport 3D~\cite{Matterport3D} & Home  & 2018 & 90     & V   & \ding{55} & \ding{55} \\
Indoor & Virtual Home~\cite{puig2018virtualhome} & Home   & 2018 & 50     & V   & \checkmark  & \checkmark \\
Indoor & Habitat~\cite{savva2019habitat}   & Home     & 2019 & -  & V   & \checkmark & \checkmark \\
Indoor & SAPIEN~\cite{xiang2020sapien}     & Home    & 2020 & 46     & V   & \checkmark & \checkmark \\
Indoor & iGibson~\cite{shen2021igibson}  & Home      & 2021 & 15     & V, L & \checkmark & \checkmark \\
Indoor & AVLEN~\cite{paul2022avlen}    & Home    & 2022 & 85     & V, T, A & \checkmark & \checkmark \\
Indoor & ProcTHOR~\cite{deitke2022️}   & Home     & 2022 & Arbitrary     & V & \checkmark & \checkmark \\
Indoor & Holodeck~\cite{yang2024holodeck}   & Home     & 2024 & Arbitrary     & V & \checkmark & \checkmark \\
Indoor & AnyHome~\cite{fu2025anyhome} & Home      & 2024 & Arbitrary     & V & \checkmark & \checkmark \\
Indoor & LEGENT~\cite{cheng2024legent}   & Home     & 2024 & Arbitrary     & V, T & \checkmark & \checkmark \\
Indoor & TDW~\cite{gan2020threedworld}    &  Home  & 2021 & -      & V, A & \checkmark & \checkmark \\
In \& Outdoor & GRUTopia~\cite{wang2024grutopia}  &  Home, City         & 2024 & 100k      & V, T & \checkmark & \checkmark \\
Outdoor & MineDOJO~\cite{fan2022minedojo}    &     Game    & 2022 & -      & V & \ding{55} & \ding{55} \\
Outdoor & MetaUrban~\cite{wu2024metaurban}    &     City    & 2024 & 13800      & V, L & \checkmark & \checkmark \\
Outdoor & UrbanWorld~\cite{shang2024urbanworld}    &    City Building    & 2024 & Arbitrary      & V & \checkmark & \checkmark \\
Outdoor & EmbodiedCity~\cite{gao2024embodiedcity}    &       City  & 2024 &    87.1k   & V, T & \checkmark & \checkmark \\
Dynamic &  UniSim~\cite{yanglearning}     &      Home, City, Simulation & 2023 & Arbitrary      & V, T & \ding{55} & \ding{55} \\
Dynamic & Streetscapes~\cite{deng2024streetscapes}   &        Street View  & 2024 & Arbitrary      & V, T & \ding{55} & \ding{55} \\
Dynamic & AVID~\cite{rigter2024avid}    &  Home, Game       & 2024 & Arbitrary      & V, T & \ding{55} & \ding{55} \\
Dynamic & EVA~\cite{chi2024eva}     &   Home, Simulation     & 2024 & Arbitrary      & V, T & \ding{55} & \ding{55} \\
Dynamic & Pandora~\cite{xiang2024pandora}    &    Home, Game, Simulation, Street View     & 2024 & Arbitrary      & V, T & \ding{55} & \ding{55} \\
Dynamic & Roboscape~\cite{shang2025roboscape}    &   Home    & 2025 & Arbitrary      & V, T & \checkmark & \ding{55} \\
Dynamic & TesserAct~\cite{zhen2025tesseract}    &   Home    & 2025 & Arbitrary      & V, T & \checkmark & \ding{55} \\
Dynamic & Aether~\cite{team2025aether}    &   Home, Street View     & 2025 & Arbitrary      & V, T & \checkmark & \ding{55} \\
Dynamic & Deepverse~\cite{chen2025deepverse}    &   Home, Simulation, Street View     & 2025 & Arbitrary      & V, T & \checkmark & \ding{55} \\
\hline
\end{tabular}
}
\label{embody-env}
\end{table}

\subsubsection{Indoor Environments}
Indoor environments offer controlled, structured scenarios where agents can perform detailed, task-specific actions such as object manipulation, navigation, and real-time interaction with users~\cite{gao2024alexa,paul2022avlen,kolve2017ai2,shen2021igibson,Matterport3D,puig2018virtualhome,savva2019habitat,xiang2020sapien}. 
Early works on establishing indoor environments like AI2-THOR~\cite{kolve2017ai2} and Matterport 3D~\cite{Matterport3D} focus on providing only visual information. These works build indoor environments by providing photorealistic settings where agents can practice visual navigation and engage in interactive tasks that mimic real-life home activities. These environments emphasize the importance of using visual-based reinforcement learning techniques that allow agents to optimize their decision-making based on environmental cues. By simulating real-world tasks like cooking or cleaning, these platforms assess an agent's capacity to generalize learned behaviors across different types of spaces and objects.
A line of further works contributes toward expanding the data modalities of the provided environments. Among these, iGibson~\cite{shen2021igibson} introduces Lidar observation as additional signal feedback, contributing to more accurate environment perception of agents.
AVLEN~\cite{paul2022avlen} further supplements audio signals allowing agents to execute tasks such as object manipulation and navigation in household-like settings. The challenge here lies in enabling agents to understand and act on multimodal input including vision, language, and sound within a constrained space.
Adding a social dimension, environments like GRUtopia~\cite{wang2024grutopia} introduce agents to spaces where they must navigate and interact with both objects and NPCs. Here, agents need to understand social dynamics, such as positioning and task sharing, which requires more advanced forms of interaction modeling. The inclusion of social interaction modules in these settings demonstrates how agents can be trained to balance human-like social behaviors with task performance.
More recently, with the development of LLMs, some works~\cite{cheng2024legent,yang2024holodeck,fu2025anyhome} seek to provide a flexible environment generation pipeline, supporting the generation of arbitrary indoor environments with language instructions.

\subsubsection{Outdoor Environments}
In contrast to indoor environments, creating outdoor environments~\cite{wang2024grutopia,gan2020threedworld,wu2024metaurban,shang2024urbanworld,fan2022minedojo} faces greater challenges due to their larger scale and increased variability. Some existing works focus on urban environments, such as MetaUrban~\cite{wu2024metaurban}, where agents are deployed to navigate in large-scale urban environments, where they encounter challenges like dynamically changing traffic, varied building structures, and social interactions with other entities. These tasks often require the use of context-aware navigation algorithms that allow agents to adjust their trajectories and behaviors based on the layout and conditions of the environment. 
However, the environments in MetaUrban are created by retrieving and organizing 3D assets from existing libraries.
Recently, utilizing advanced generative techniques, UrbanWorld~\cite{shang2024urbanworld} significantly enhances the scope of outdoor environments, using 3D generative models to create complex, customizable urban spaces that allow for more diverse urban scenarios. This shift from static asset-based environments to generative ones ensures that agents are exposed to a wider variety of tasks, from navigating unfamiliar street layouts to interacting with new types of objects or structures. 
In addition to the above real open-world generation works, there are also some virtual open-world platforms like MineDOJO~\cite{fan2022minedojo} that extend these challenges even further by simulating procedurally generated, sandbox-like environments. These platforms, inspired by the open-ended world of Minecraft, push agents to engage in tasks like resource collection, construction, and survival, demanding continuous exploration and adaptive learning. In such environments, agents are motivated to seek out new information and adapt their behavior to finish given tasks. Training in such environments can help agents learn knowledge across a broad range of tasks and terrains, enabling them to operate effectively in various outdoor environments.

\subsubsection{Dynamic Environments}
Dynamic environments mark a significant evolution from traditional, static simulators by utilizing generative models to create flexible, real-time simulations. Unlike predefined environments that require manual adjustments, these models allow for the dynamic creation of a wide variety of scenarios, enabling agents to experience diverse, first-person perspectives. This shift provides agents with richer, more varied training experiences, improving their adaptability and generalization in complex, unpredictable real-world situations.
A representative work is UniSim~\cite{yanglearning}, which dynamically generates robot manipulation video sequences based on input conditions like spatial movements, textual commands, and camera parameters. Leveraging multimodal data from 3D simulations, real-world robot actions, and internet media, this system generates varied, realistic environments where agents can practice tasks like object manipulation and navigation. The key advantage of this approach is its flexibility, allowing agents to adapt to various scenarios without the limitations of static physical environments.
Pandora~\cite{xiang2024pandora} expands the dynamic environment generation from robot actions in Unisim to wider domains including human and robot actions in both indoor and outdoor scenes.
Another subsequent work, AVID~\cite{rigter2024avid} builds on UniSim by conditioning on actions and modifying noise predictions from a pre-trained diffusion model to generate action-driven visual sequences for dynamic environment generation.
Beyond the video diffusion-based framework of Unisim, EVA~\cite{chi2024eva} introduces an additional vision-language model for embodied video anticipation, producing more consistent embodied video predictions.
As for the generation of open-world dynamic environments, Streetscapes~\cite{deng2024streetscapes} employs autoregressive video diffusion models to simulate urban environments where agents must navigate dynamic challenges like changing weather and traffic. These environments offer consistently coherent, yet flexible, urban settings, exposing agents to real-world-like variability.
The core trend in dynamic environments is the use of generative world models that provide scalable, adaptable simulations. This approach significantly reduces the manual effort required for environment setup, allowing agents to train across a diverse range of scenarios quickly. Moreover, the focus on first-person training closely mimics real-world decision-making, enhancing the agents' ability to adapt to evolving situations. These advances are key in developing embodied environments supporting agent learning in complex, dynamic scenarios.

Given the above developments, it is evident that world models as embodied environments have made significant advances in simulating the environmental transition of the real world. Current research predominantly focuses on developing indoor, static environments, with notable efforts expanding to large-scale outdoor and dynamic simulation environments. A promising direction is to construct dynamic environments, which can provide first-person, action-conditioned future world prediction, enabling agents to better adapt to unseen conditions. Simultaneously, recent advancements in building dynamic embodied worlds emphasize the integration of physical constraints. For instance, Aether~\cite{team2025aether} enhances geometric knowledge learning by employing camera trajectories as action-driven RGB-D video generation. TesserAct~\cite{zhen2025tesseract} further incorporates normal maps as physical constraints for video generation. Roboscape~\cite{shang2025roboscape} integrates depth maps and keypoint dynamics during video generation to learn and produce more realistic motions and spatial structures. Additionally, Deepverse~\cite{chen2025deepverse} proposes incorporating geometric predictions from previous timesteps into current predictions conditioned on actions. These methods collectively enhance the realism and physical adherence of the dynamic world, thereby creating more authentic and reliable simulated environments for embodied agents.

\section{Application Domains} \label{sec:application}

\subsection{Game Intelligence}

Gaming environments represent an ideal testbed for world model research, offering controlled yet complex domains that demand a sophisticated understanding of physics, causality, and interactive dynamics. Unlike real-world applications where ground truth is often ambiguous or inaccessible, games provide well-defined rule systems and clear action-consequence relationships that enable precise evaluation of world model capabilities.

More importantly, world model technologies are fundamentally transforming game development and player experiences in unprecedented ways. Traditional game development relies on manually coded rules, pre-designed assets, and scripted interactions that constrain creative possibilities and require extensive development resources. World models offer a paradigm shift toward generative game systems that can autonomously create new content, adapt dynamically to player behavior, and enable emergent gameplay experiences that were previously impossible to achieve through conventional programming approaches.

Recent developments demonstrate three critical capability dimensions in gaming applications:

\textbf{Interactivity.} The ability to respond appropriately to user inputs represents a fundamental requirement for gaming world models. GameNGen demonstrates this capability by creating a fully neural game engine that enables real-time interaction with complex environments, running at 20 frames per second while maintaining stable gameplay over extended sessions~\cite{valevski2024diffusion}. Similarly, GameGen-X introduces a specialized module to incorporate game-related multi-modal control signals, unifying character interaction and scene content control for the first time in video generation~\cite{che2024gamegen}. Matrix-Game advances this further by training an over 17 billion parameter model capable of precise control over both character actions and camera movements through fine-grained keyboard and mouse action annotations~\cite{zhang2025matrix}.

\textbf{Consistency.} Maintaining coherent game states across temporal sequences poses significant challenges for generative models. Recent research tackles both numerical consistency (ensuring gameplay mechanics correctly reflect score changes and quantitative elements) and spatial consistency (preventing jarring scene transitions) in games~\cite{chen2025model}. MineWorld addresses consistency through visual-action autoregressive transformers that learn rich representations of game states and action-state relationships simultaneously~\cite{guo2025mineworld}. The model's parallel decoding algorithm enables real-time generation while maintaining temporal coherence across extended gameplay sequences. WHAM (World and Human Action Model) further exemplifies this advancement by generating consistent and diverse gameplay sequences while persisting user modifications—capabilities identified as critical for supporting creative practices in game development~\cite{kanervisto2025world}.

\textbf{Generalization Across Diverse Environments.} The ability to adapt across varied gaming scenarios and environments represents perhaps the most challenging dimension. GameFactory addresses this through scene-generalizable action control, leveraging open-domain generative priors from pre-trained video diffusion models to create entirely new games beyond fixed styles and scenes~\cite{yu2025gamefactory}. Recent work further creates a "generative infinite game" that transcends traditional finite, hard-coded systems~\cite{li2024unbounded}. Using specialized distilled LLMs for dynamic game mechanic generation and dynamic regional image prompt adapters for consistent visual generation, the system enables open-ended mechanics that can emerge naturally from the underlying generative models. The exploration-driven approach in virtual environments demonstrates another path toward generalization, where exploration agents rely entirely on world model uncertainty to deliver diverse training data, adapting easily to new environments without environment-specific rewards~\cite{savov2025exploration}.

\subsection {Embodied Intelligence}
Embodied intelligence focuses on creating agents that can perceive, understand, and interact effectively with the complex physical world. A central challenge in this domain is equipping robots with the ability to reason about their environment's dynamics to support robust, real-time decision-making. World models have emerged as a transformative paradigm that directly addresses this need, empowering robots with the critical ability to perceive, predict, and act effectively. This progress is driven in part by advances in neural architectures \cite{vaswani2017attention, ho2020denoising} and learning algorithms \cite{schulman2017proximal, rafailov2024direct}, which enable robots to build implicit representations that capture key aspects of the external world. Complementarily, prediction models \cite{finn2016unsupervised, finn2017deep} offer the ability to forecast future environmental states, moving beyond static abstractions to support anticipatory and adaptive behavior. Together, these capabilities make it increasingly feasible for robots to learn directly from real-world interactions. In Table~\ref{tab:robot_wm}, we summarize the core learning tasks involved in constructing world models for robotics, categorized according to the three major perspectives outlined above (typical examples shown in Figure ~\ref{fig:robot_wm}).

\subsubsection{Learning Implicit Representation}

Traditional robotic tasks (e.g., object grasping) are typically performed in highly structured environments where the critical components are explicitly modeled~\cite{kleeberger2020survey,durrant2006simultaneous}, eliminating the need for the robot to independently learn or adapt its understanding of the world.
However, when the robot is deployed in unfamiliar environments, especially those in which key features or dynamics have not been explicitly modeled, tasks that were previously successful may fail as the robot struggles to generalize to these unknown features~\cite{mnih2015human,kahn2017uncertainty}.
Thus, enabling a robot to learn an implicit representation of its environment is a crucial first step toward achieving intelligence.

To help a robot understand the objects in the world, visual models such as convolutional neural networks (CNNs)~\cite{lecun1998gradient,krizhevsky2012imagenet,girshick2014rich} and vision transformers (ViT)~\cite{dosovitskiy2020image,wang2024repvit} integrate visual characteristics of entities into representations, making it possible for robots to recognize critical objects for tasks.
RoboCraft~\cite{shi2024robocraft} transfers visual observation into particles and captures the structure of the underlying system through a graph neural network.
Moreover, other attempts are made for the sensing of physical 3D space.
PointNet~\cite{qi2017pointnet,qian2022pointnext} first encodes the unstructured 3D point clouds with asymmetrical functions, capturing the spatial characteristics of the environment.
A recent work~\cite{gornet2024automated} assembles observations acquired along local exploratory paths into a global representation of the physical space within its latent space, enabling robots to tail and approach specific targets.
SpatialLM~\cite{spatiallm} further advances this direction by processing raw 3D point clouds into structured 3D scene representations with semantic labels, enhancing spatial reasoning for complex tasks in robotics and autonomous driving.
With the advancement of language comprehension in LLMs~\cite{touvron2023llama,brown2020language,du2022glam}, a novel paradigm for enabling robots to capture task intentions involves describing the task in textual form and then obtaining a textual representation through LLMs~\cite{mu2023clarifygpt,gestrin2024nl2plan,hua2024gensim2,wang2023gensim}.
BC-Z~\cite{jang2022bc} utilizes language representations as task representations, enhancing the multi-task performance of robots.
Text2Motion~\cite{lin2023text2motion} splits the natural language instruction into task-level and motion-level plans with LLM to handle complex sequential manipulation tasks.

\begin{table}[!t]
\footnotesize
\renewcommand{\arraystretch}{0.9}
\caption{Core learning tasks involved in constructing world models for robotics.}
\vspace{-1em}
\centering
\begin{tabular}{c|c|c|c|c|c}
  \toprule
  & Task & Model & Year & Input & Backbone \\
\midrule
  \multirow{9}{*}{\makecell{Learning \\ Inner \\ Representation}} 
  & \multirow{3}{*}{\makecell{Visual \\ Representation}} 
  & CNN~\cite{lecun1998gradient} & 1998 & Image & CNN \\
  & & ViT~\cite{dosovitskiy2020image} & 2020 & Image & Transformer \\
  & & RoboCraft~\cite{shi2024robocraft} & 2024 & Image & GNN \\
  \cmidrule(l){2-6}
  & \multirow{3}{*}{\makecell{3D \\ Representation}} 
  & PointNet~\cite{qi2017pointnet} & 2017 & 3D point clouds & MLP \\
  & & Predictive Coding~\cite{gornet2024automated} & 2024 & Image & ResNet \\
  & & SpatialLM~\cite{spatiallm} & 2025 & 3D point clouds & MLLM \\
  \cmidrule(l){2-6}
  & \multirow{3}{*}{\makecell{Task \\ Representation}} 
  & BC-Z~\cite{jang2022bc} & 2022 & Text\& Video & LLM\& ResNet \\
  & & Text2Motion~\cite{lin2023text2motion} & 2023 & Text & LLM \\
  & & Gensim~\cite{wang2023gensim} & 2023 & Text & LLM \\
\midrule
  \multirow{3}{*}{\makecell{\\ \\ \\  Predicting \\ Future \\ Environment}} 
  & \multirow{4}{*}{\makecell{\\ \\ \\ \\ Video \\ Prediction}} 
  & UniPi~\cite{du2024learning} & 2024 & Video & Diffusion \\
  & & VIPER~\cite{escontrela2024video} & 2024 & Video & Transformer \\
  & & GR-2~\cite{cheang2024gr} & 2024 & Text \& Video & Transformer \\
  & & IRASim~\cite{zhu2024irasim} & 2024 & Trajectory & Diffusion \\
  & & VPP~\cite{hu2024video} & 2024 & Text & Diffusion \\
  & & DreamGen~\cite{jang2025dreamgen} & 2025 & Text & Diffusion \\
  & & Roboscape~\cite{shang2025roboscape} & 2025 & Trajectory & Transformer \\
  & & EVAC~\cite{jiang2025enerverse} & 2025 & Trajectory & Diffusion \\
  & & Genie Envisioner~\cite{liao2025genie} & 2025 & Text & Diffusion \\
  & & Vidar~\cite{feng2025generalist} & 2025 & Text & Diffusion \\
  & & V-JEPA 2~\cite{assran2025v} & 2025 & Trajectory & Transformer \\
\midrule
  \multirow{4}{*}{\makecell{Real-world \\ Planning}} 
  & \multirow{3}{*}{\makecell{Real-World \\ Adaptation}} 
  & DayDreamer~\cite{wu2023daydreamer} & 2023 & Video & RSSM \\
  & & SWIM~\cite{mendonca2023structured} & 2023 & Video & Transfer Learning \\
  & & CoSTAR~\cite{agha2021nebula} & 2021 & Multimodal & Belief Space \\
  \cmidrule(l){2-6}
  & \multirow{1}{*}{\makecell{Evaluation}} 
 & OpenEQA~\cite{majumdar2024openeqa} & 2024 & Image\& Text & LLM \\
  \bottomrule
\end{tabular}
\label{tab:robot_wm}
\vspace{-2em}
\end{table}

\subsubsection{Predicting Future States of the Environment}

World models are at the forefront of robotic research, primarily enabling advancements across three application areas: synthetic data generation, action guidance through imagined futures, and environment simulation for policy evaluation.

First, embodied world models can synthesize high-quality robot action videos to augment real-world collected data, thereby enhancing the training of downstream robotic policy models. For instance, DreamGen~\cite{jang2025dreamgen} proposes a four-stage pipeline to generate neural trajectories, a form of synthetic robot data derived from video world models. This approach significantly boosts robotic operation success and generalization in Vision-Language-Action (VLA) model training, particularly for contact-rich tasks. Roboscape~\cite{shang2025roboscape} integrates physical laws during video generation, leading to synthetic data with improved motion plausibility and spatial accuracy, which in turn yields substantial gains when incorporated into VLA training. EVAC~\cite{jiang2025enerverse} improves generalization by expanding training data with diverse failure trajectories, utilizing a multi-level action-conditioning mechanism and ray map encoding for dynamic multi-view image generation, effectively serving as both a data engine and an evaluator by augmenting human-collected trajectories.

Second, embodied world models guide robot action generation by leveraging imagined future observations. A key recent insight is the use of generative video models—particularly those leveraging diffusion~\cite{esser2023structure,chi2023diffusion,black2023zero,helearning} and transformer architectures~\cite{yu2023magvit,yan2021videogpt}—to implicitly learn environmental dynamics directly from visual data. For instance, UniPi~\cite{du2024learning} explicitly frames action prediction as a video generation problem, conditioning a constrained diffusion model on the current state to visualize future scenarios. Similarly, VIPER~\cite{escontrela2024video} employs a pretrained autoregressive transformer to guide robotic actions, effectively leveraging rich representations learned from expert demonstration videos. Moreover, models like GR-2~\cite{cheang2024gr} benefit from the vast scale of internet videos to establish robust priors, subsequently fine-tuning on specific robotics tasks to generate accurate image predictions and action trajectories. VPP~\cite{hu2024video} learns robot actions by fine-tuning a video generation model based on text instructions, subsequently deriving actions from an inverse dynamics model conditioned on visual representations. Similarly, Genie Envisioner~\cite{liao2025genie} leverages a pretrained embodied video generation foundation model, connected with a lightweight parallel flow-matching action model that translates language-conditioned visual latent features into fine-grained, low-latency motor commands. Vidar~\cite{feng2025generalist} proposes a two-stage framework for robotic action prediction, combining large-scale, diffusion-based video pre-training with a novel masked inverse dynamics model. Distinctly, V-JEPA 2~\cite{assran2025v} models world state transitions in a latent space and performs action planning through Model Predictive Control (MPC), which involves extensive sampling of possible action trajectories and selecting the optimal one based on energy optimization.

Third, embodied world models can function as environment simulators for policy evaluation. IRASim~\cite{zhu2024irasim} and Roboscape~\cite{shang2025roboscape} both leverage world models for trajectory-to-video generation tasks, starting from an initial given frame. Their demonstrated high correlation between policy model evaluation in the world model and in real environments indicates that the learned world models accurately capture world transition dynamics. GE-Sim~\cite{liao2025genie} also exemplifies this by designing a closed-loop interaction between a policy and a world model, enabling scalable and flexible simulation without requiring manual environment modeling.

Collectively, these methods demonstrate the profound promise of generative, vision-centric world modeling as a foundation for anticipatory robotic control and simulation. This significantly enhances robots’ capabilities to reason about future states and improve long-term task performance.

\subsubsection{From Simulation to Real World}
Deep reinforcement learning has demonstrated remarkable capabilities in robotics, enabling autonomous performance in complex tasks such as stable locomotion~\cite{smith2022walk,kumar2021rma}, precise object manipulation~\cite{yu2022se,dogar2019multi}, and intricate activities like tying shoelaces~\cite{aldaco2024aloha}. However, its practicality remains significantly limited by low sample efficiency. For instance, training a robot to solve a Rubik’s Cube in the real world can require tens of thousands of simulated years~\cite{akkaya2019solving}. Consequently, most robot training is conducted within simulation environments, leveraging distributed training techniques to enhance efficiency~\cite{rudin2022learning,ha2020learning}. Unfortunately, due to discrepancies between simulations and real-world conditions, policies trained in simulation often fail when directly transferred to physical robots, particularly in complex or novel environments.

A pivotal recent insight is that world models can effectively bridge this simulation-to-reality gap by learning generalized representations of real-world dynamics. For example, NeBula~\cite{agha2021nebula} constructs a structured belief space that enables reasoning and rapid adaptation across diverse robot morphologies and unstructured environments. DayDreamer~\cite{wu2023daydreamer} further demonstrates the capability of generalized world models, allowing robots to directly learn locomotion in real-world environments within hours, significantly reducing the reliance on extensive simulations. Additionally, SWIM~\cite{mendonca2023structured} highlights the power of human-video-based learning combined with minimal real-world fine-tuning, enabling task generalization with less than 30 minutes of interaction. These examples illustrate that by building robust, real-world-oriented internal representations, world models substantially narrow the gap between simulation and reality, facilitating rapid adaptation and generalization in robotics.

\subsection{Urban Intelligence}

\subsubsection{Autonomous Driving}
In recent years, with the rapid advancement of vision-based generative models~\cite{ho2020denoising, song2020score, videoworldsimulators2024} and multimodal large language models~\cite{liu2023llava, achiam2023gpt4}, world models have attracted growing interest in the field of autonomous driving. 
The modern autonomous driving pipeline is typically divided into four key components: \textit{perception}, \textit{prediction}, \textit{planning}, and \textit{control}. Among these, the perception and prediction stages correspond to driving scene understanding—i.e., learning an implicit representation of the vehicle’s external environment. In parallel, recent surveys~\cite{guan2024world} highlight the emergence of end-to-end world simulators that learn to simulate realistic driving environments based on multimodal inputs—such as images, point clouds, trajectories, and language—and then generate future states to support downstream tasks like planning and decision-making. These two perspectives align well with our earlier categorization of world models, and in the following, we detail their applications and developments within the autonomous driving domain accordingly.

\begin{figure}[htbp]
    \centering
    \includegraphics[width=0.9\linewidth]{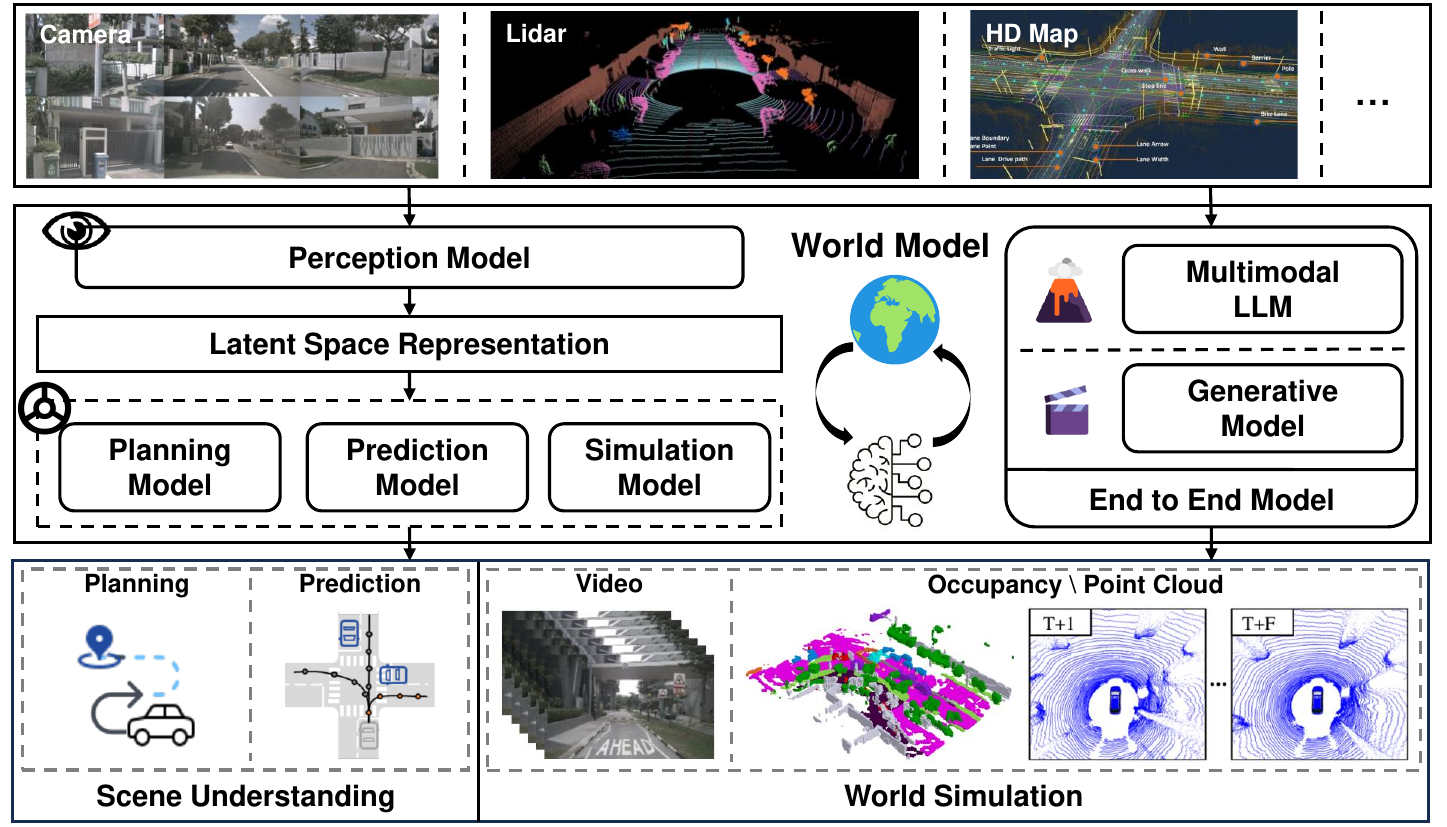}
    \caption{Application of world model in autonomous driving.}
    \label{fig:autonomous}
    \vspace{-1em}
\end{figure}

\textit{Learning Implicit Representations.}
Autonomous vehicles typically utilize cameras, radar, and lidar to perceive the real world, gathering information through images, video data, and point cloud data. In the initial decision-making paradigm~\cite{chen2019modelfreedeepreinforcementlearning, Saxena_2020}, models often take perceptual data as input and directly output motion planning results for the autonomous vehicle. Conversely, when humans operate vehicles, they typically observe and predict the current and future states of other traffic participants to determine their own driving strategies~\cite{survey_trajpre_2022}. Thus, learning the implicit representation of the world through perceptual data and predicting the future states of the surrounding environment is a crucial step in enhancing the decision-making reliability of autonomous vehicles. We consider this process as it manifests in how autonomous vehicles learn a world model in latent space.

As shown in the left half of Figure~\ref{fig:autonomous}, before the advent of multimodal large models and end-to-end autonomous driving technologies~\cite{hu2023planningorientedautonomousdriving}, the perception and prediction tasks of autonomous vehicles were typically assigned to distinct modules, each trained on their respective tasks and datasets. The perception module processed data from images, point clouds, and other sources to accomplish tasks such as object detection and map segmentation, projecting the perceived world into an abstract geometric space. Furthermore, the prediction module would typically operate within these geometric spaces to forecast the future states of the surrounding environment, including the trajectories and motions of traffic participants.

\begin{table}[t]
\footnotesize
\renewcommand{\arraystretch}{0.9}
\caption{Comparison of existing works in scene understanding and world simulation.}
\vspace{-1em}
\centering
\resizebox{\textwidth}{!}{
\begin{tabular}{c|c|c|c|c|c|c}
  \toprule
  & Task  & Work & Year & Data Modality & Technique & Task Description \\
\midrule
  \multirow{15}{*}{\makecell{Driving Scene \\ Understanding}} 
  & \multirow{7}{*}{\makecell{Perception}}
  & Faster r-cnn~\cite{NIPS2015_14bfa6bb} & 2015 & Camera & CNN & Obeject Detection \\
  & &Pointnet~\cite{qi2017pointnetdeephierarchicalfeature} & 2017 & Lidar & MLP & 3D Classification \\
  & &MultiNet~\cite{teichmann2018multinetrealtimejointsemantic} & 2018 & Camera & CNN & Semantic Segmentation \\
  & &OmniDet~\cite{kumar2023omnidetsurroundviewcameras} & 2021 & Camera & CNN \& Attention & Multi-task Visual Perception\\
  & &YOLOP~\cite{yolop_2022} & 2022 & Camera & CNN & Object Detection \\
  & &BEVFormer~\cite{li2022bevformer} & 2022 & Camera & Attention & 3D Visual Perception \\
  & &Transfusion~\cite{bai2022transfusion} & 2022 & Camera \& Lidar & Transformer & 3D Object Detection \\
  \cmidrule(l){2-7}
  & \multirow{5}{*}{\makecell{Prediction}} 
  & Wayformer~\cite{nayakanti2022wayformermotionforecastingsimple} & 2022 & Geometric Space & Attention & Trajectory Prediction \\
  & & MTR~\cite{shi2022motion} & 2022 & Geometric Space & Transformer & Trajectory Prediction \\
  & & QCNet~\cite{zhou2023query} & 2023 & Geometric Space & Transformer & Trajectory Prediction \\
  & & HPTR~\cite{zhang2023realtimemotionpredictionheterogeneous} & 2023 & Geometric Space & Transformer & Trajectory Prediction \\
  & & Jiang \textit{et al.}~\cite{jiang2023motiondiffusercontrollablemultiagentmotion} & 2023 & Geometric Space & Diffusion & Trajectory Prediction \\
  \cmidrule(l){2-7}
  & \multirow{3}{*}{\makecell{End to End\\Scene\\Understanding}} 
  & UniAD~\cite{hu2023planningorientedautonomousdriving} & 2023 & Camera & Transformer & Motion Planning \\
  & & TOKEN~\cite{tian2024tokenizeworldobjectlevelknowledge} & 2024 & Camera & MLLM & Motion Planning \\
  & & OmniDrive~\cite{kumar2023omnidetsurroundviewcameras} & 2024 & Camera & MLLM & Motion Planning \\
\midrule
  \multirow{11}{*}{\makecell{Driving World \\ Simulation}}
  & \multirow{4}{*}{\makecell{Motion\\Simulation}}
  & SUMO~\cite{SUMO2018} & 2000 & Geometric Space & Rule-based & Traffic Simulation \\
  & & Metadrive~\cite{li2022metadrive} & 2022 & Geometric Space & Data-driven & Traffic Simulation \\
  & & Trafficbots~\cite{zhang2023trafficbots} & 2023 & Geometric Space & Transformer & Traffic Simulation\\
  & & Waymax~\cite{gulino2024waymax} & 2024 & Geometric Space & Data-driven & Traffic Simulation \\
  \cmidrule(l){2-7}
  & \multirow{7}{*}{\makecell{End to End \\ Sensor \\Simulation}}
  & GAIA-1~\cite{hu2023gaia1generativeworldmodel} & 2023 & Camera & Transformer & Video Generation \\
  & &DriveDreamer~\cite{wang2023drivedreamerrealworlddrivenworldmodels} & 2023 & Camera & Diffusion & Video Generation \\
  & &Drive-WM~\cite{wang2023drivingfuturemultiviewvisual} & 2023 & Camera & Diffusion & Video Generation \\
  & &OccWorld~\cite{zheng2023occworld} & 2023 & Occupancy & Attention & Occupancy Generation \\
  & &OccSora~\cite{wang2024occsora} & 2024 & Occupancy & Diffusion & Occupancy Generation\\
  & &Vista~\cite{gao2024vista} & 2024 & Camera & Diffusion & Video Generation \\
  & &Copilot4D~\cite{zhang2024copilot4dlearningunsupervisedworld} & 2024 & Lidar & Diffusion & Point Cloud Generation\\
  \bottomrule
\end{tabular}}
\label{tab:autonomous}
\vspace{-2em}
\end{table}

The processing of perceptual data is closely tied to the evolution of deep learning technologies, as shown in Table~\ref{tab:autonomous}. Pointnet~\cite{qi2017pointnetdeephierarchicalfeature}, introduced in 2017, was the first to employ deep learning methods for processing point cloud data. As convolutional neural networks advanced, perception techniques based on image data, exemplified by YOLOP~\cite{yolop_2022} and MultiNet~\cite{teichmann2018multinetrealtimejointsemantic}, emerged and excelled in driving scene understanding tasks~\cite{He2019Mgnet, vu2022hybridnetsendtoendperceptionnetwork, kumar2023omnidetsurroundviewcameras, zhou2022matrixvtefficientmulticamerabev}. In recent years, the transformer architecture has gained prominence in natural language processing, and this technology has also been applied to image data understanding. BEVFormer~\cite{li2022bevformer} utilizes the attention mechanism to integrate images from multiple camera angles, constructing an abstract geometric space from a bird's-eye view, and achieving state-of-the-art results in various tasks, including object detection. Additionally, Transfusion~\cite{bai2022transfusion} enhances perceptual accuracy by fusing lidar and camera data through a cross-attention approach. Building on the perceptual results, a series of techniques such as RNNs~\cite{altche2017lstm, zyner2019naturalistic, kawasaki2020multimodal}, CNNs~\cite{phan2020covernet, cui2019cnn, chou2020predicting}, and Transformers~\cite{huang2022multi, ngiam2021scene, shi2022motion, zhou2023query} have been employed to encode historical scene information and predict the future behaviors of traffic participants.

With the emergence and rapid development of multimodal large language models in recent years, many efforts have sought to apply the general scene understanding capabilities of these models to the field of autonomous driving. TOKEN~\cite{tian2024tokenizeworldobjectlevelknowledge} tokenizes the whole traffic scene into object-level knowledge, using the reasoning ability of the language model to handle the long-tail prediction and planning problems, OmniDrive~\cite{kumar2023omnidetsurroundviewcameras} sets up llm-based agents and covers multiple tasks including scene description, counterfactual reasoning and decision making through visual question-answering.

\textit{World Simulators.}
As shown in Table~\ref{tab:autonomous}, before the emergence of multimodal large models and vision-based generative models, traffic scenario simulations are often conducted in geometric spaces. The scene data on which these simulations rely is typically collected by the perception modules of autonomous vehicles or constructed manually. These simulations represent future states of the scenario in the form of geometric trajectories~\cite{SUMO2018, li2022metadrive, gulino2024waymax, zhang2023trafficbots}, which require further modeling and rendering to produce outputs suitable for vehicle perception. The cascading of multiple modules often results in information loss and increases the complexity of simulations, making scenario control more challenging. Furthermore, realistic scene rendering typically requires substantial computational resources, which limits the efficiency of virtual traffic scenario generation. 

Using diffusion-based video generation models as a world model partially addresses the aforementioned issues. By training on large-scale traffic scenario datasets, diffusion models can directly generate camera perception data that closely resembles reality. Additionally, the inherent controllability of diffusion models, combined with text-image alignment methods like CLIP~\cite{radford2021learningtransferablevisualmodels}, enables users to exert control over scenario generation in a straightforward manner. The GAIA-1~\cite{hu2023gaia1generativeworldmodel} and DriveDreamer series ~\cite{wang2023drivedreamerrealworlddrivenworldmodels, zhao2024drivedreamer2llmenhancedworldmodels} are among the first to employ this method for constructing world models. Building on this foundation, Drive-WM~\cite{wang2023drivingfuturemultiviewvisual} introduces closed-loop control for planning tasks, while Vista~\cite{gao2024vista} focuses on improving the resolution of generated results and extending prediction duration. In addition to methods that predict future states in video space, many other works have explored different forms of vehicle perception data. OccWorld~\cite{zheng2023occworld} and OccSora~\cite{wang2024occsora} predict the future state of the world by forecasting 3D occupancy grids, whereas Copilot4D~\cite{zhang2024copilot4dlearningunsupervisedworld} constructs a world model by predicting changes in radar point cloud data. Compared to video data, these types of features better reflect the spatial characteristics of traffic scenarios.

\begin{table}[ht]
\centering
\caption{Applications of World Models in Autonomous Logistics and Urban Analytics}
\label{tab:urban_world_models_updated}
\resizebox{0.8\textwidth}{!}{
\begin{tabular}{@{}lllll@{}}
\toprule
\textbf{Category}            & \textbf{Sub-category} & \textbf{Paper}        & \textbf{Year} & \textbf{Venue} \\ \midrule
\multirow{8}{*}{Autonomous Logistics} & \multirow{4}{*}{Micromobility} & NWM~\cite{bar2025navigation}          & 2025                     & CVPR                               \\
                                    &                              & URBAN-SIM~\cite{wu2025towards} & 2025                     & CVPR                               \\
                                    &                              & Vid2Sim~\cite{xie2025vid2sim}                          & 2025                     & CVPR                               \\
                                    &                              & CityWalker~\cite{liu2025citywalker}                       & 2025                     & CVPR                               \\ \cmidrule(l){2-5} 
                                    & \multirow{4}{*}{Aerial}      & AirScape~\cite{zhao2025airscape}                         & 2025                     & ACM MM                             \\
                                    &                              & CityNavAgent~\cite{zhang2025citynavagent}                     & 2025                     & ACL                                \\
                                    &                              & CityEQA~\cite{zhao2025cityeqa}                          & 2025                     & EMNLP                              \\
                                    &                              & UrbanVideo-Bench~\cite{zhao2025urbanvideo}                 & 2025                     & ACL                                \\ \midrule
\multirow{8}{*}{Urban Analytics}      & \multirow{3}{*}{Knowledge}   & CityGPT~\cite{feng2024citygpt}                          & 2025                     & KDD                                \\
                                    &                              & UrbanLLaVA~\cite{feng2025urbanllava}                       & 2025                     & ICCV                               \\
                                    &                              & GeoLLM~\cite{manvi2023geollm}                           & 2024                     & ICLR                               \\ \cmidrule(l){2-5} 
                                    & \multirow{2}{*}{Prediction}  & GPS-to-Image~\cite{feng2025gps}          & 2025                     & CVPR                               \\
                                    &                              & AI's Blind Spots~\cite{beneduce2025ai}                 & 2025                     & arXiv                              \\ \cmidrule(l){2-5} 
                                    & \multirow{3}{*}{Understanding} & AgentMove~\cite{feng2025agentmove}                        & 2025                     & NAACL                              \\
                                    &                              & CAMS~\cite{du2025cams}                             & 2025                     & arXiv                              \\
                                    &                              & PIGEON~\cite{lan2025open}  & 2025                     & ACL                                \\ \bottomrule
\end{tabular}}
\end{table}

\subsubsection{Autonomous Logistics}

This section introduces the application of world models in the autonomous logistics for urban scenario, primarily focusing on two areas: 1) miniature mobile logistics vehicles and 2) low-altitude aerial vehicles. In the research of both domains, we will introduce specific work from the perspectives of both understanding and prediction, which are shown in Table~\ref{tab:urban_world_models_updated}.

In the context of miniature mobile logistics vehicles, which represent an extension of autonomous driving into embodied intelligence scenarios, there is a need to cope with more complex surrounding environments and human-robot interactions. The core task is navigation, which involves comprehending the surrounding world to achieve more efficient and safer movement and interaction. In terms of understanding, studies like Vid2Sim~\cite{xie2025vid2sim} and CityWalker~\cite{liu2025citywalker} leverage the vast amount of videos available on the internet to understand diverse environments and interaction scenarios among movement. This understanding is then used to train the robot's navigation policy, aiming for strong generalization and controllability. In terms of predicting the future, two typical paradigms exist. The first involves constructing a realistic physical simulator~\cite{dosovitskiy2017carla,wu2025towards} to generate diverse scenarios. Through large-scale training in this virtual environment, the robot's capabilities are enhanced and then applied to the real world. The other paradigm is based on an interactive and controllable world model~\cite{bar2025navigation} derived from video generation models. This model generates potential future scenes that the robot might encounter based on its actions, specifically its trajectory, thereby enhancing its generalized navigation capabilities in a variety of scenarios.

Regarding low-altitude aerial vehicles, the vast majority of current work focuses on applications in understanding, particularly in scene comprehension and navigation~\cite{zhao2025cityeqa,zhao2025urbanvideo,zhang2025citynavagent}, while applications in prediction~\cite{zhao2025airscape} are still in their nascent stages. In the realm of understanding, a typical paradigm involves comprehending the current urban scene and its key elements from images~\cite{zhao2025cityeqa} or videos~\cite{zhao2025urbanvideo}. This provides sufficient support for subsequent tasks such as navigation~\cite{zhang2025citynavagent}. Understanding diverse and complex urban scenes relies on strong prior knowledge of the urban environment, especially the rich world knowledge and common sense derived from large language models~\cite{achiam2023gpt4}. In the domain of generation, AirScape~\cite{zhao2025airscape} stands as the first world model for low-altitude aerial vehicles. It enables the prediction of future scenes based on the aerial's actions while maintaining physical and spatiotemporal consistency. This provides a new approach and environment for the efficient training and task resolution of low-altitude aerial vehicles in the future.

\subsubsection{Urban Analytics}

In this section, we primarily introduce the application of world models in Urban Analytics. Building on the premise mentioned in Section 3.2 that large language models have already learned worldly geographical knowledge~\cite{feng2024citygpt,manvi2023geollm}, we will introduce relevant work from the perspectives of both understanding and prediction. Related works are summarized in Table~\ref{tab:urban_world_models_updated}.

In terms of understanding, on one hand, concerning the environment, multimodal large language models such as UrbanLLaVA~\cite{feng2025urbanllava} leverage the rich world knowledge embedded within models to perform generalizable tasks like urban scene recognition and comprehension~\cite{feng2024citybench}. On the other hand, regarding human behavior within the environment, AgentMove~\cite{feng2025agentmove} and CAMS~\cite{du2025cams} utilize existing or specifically enhanced urban geospatial knowledge within models~\cite{feng2024citygpt} to model human mobility patterns. Meanwhile, PIGEON~\cite{lan2025benchmarking} draws upon the common sense knowledge of large language models to understand daily human needs and their corresponding behaviors, thereby achieving accurate understanding and prediction even for infrequent scenarios.

In terms of prediction, GPS-to-Image~\cite{feng2025gps} attempts to use GPS signals to control the style and scene of generated images, demonstrating the feasibility of using models to grasp the relationship between geographical location and urban scenery. However, researchers~\cite{beneduce2025ai} have further discovered that existing image generation models still have significant room for improvement in accurately distinguishing the cultural styles and scenic characteristics of different geographical locations.

Overall, the application of world models in urban analytics is still relatively limited, indicating substantial potential for future applications.

\subsection {Societal Intelligence}

Societal intelligence is the collective ability of a society to sense its environment, reason about possible futures, and coordinate actions toward shared goals~\cite{levy1997collective,malone2015handbook,malone2018superminds}. It emerges from the interactions among individuals, institutions, and their surroundings~\cite{epstein2012generative}. One effective way to operationalize societal intelligence in silico is through \textit{social simulacra}~\cite{park2022social}, which are virtual social computing systems populated by diverse agents capable of realistic, human-like behaviors. Traditionally, such systems have been constructed using expert-defined rules~\cite{schelling1971dynamic,brock1998heterogeneous}, which encode domain knowledge into explicit behavior specifications, or through reinforcement learning~\cite{zheng2022ai}, which optimizes agent strategies via trial-and-error in simulated environments. While effective in certain contexts, these approaches often lead to overly simplistic dynamics or limited interpretability. The advent of LLMs offers a transformative foundation for creating richer and more convincing simulacra, enabling both the reproduction of stylized facts~\cite{li2024econagent} and the generation of credible predictions~\cite{piao2025agentsociety}.

In this survey, we examine world models within the framework of societal intelligence from two complementary perspectives. First, social simulacra can function as an explicit world model that mirrors real-world societies, providing a structured environment in which societal intelligence can emerge. Second, agents in simulacra can develop implicit world models through interaction, forming internal representations of the external environment that guide their decisions and social behaviors. These two perspectives—mirroring real-world society and understanding the external world—structure the following subsections. Table~\ref{tab:simulacra} summarizes representative works across both perspectives.

\begin{table}[t]
\footnotesize
\renewcommand{\arraystretch}{0.95}
\caption{Representative works of LLM-driven social simulacra from two perspectives: mirroring real-world society and understanding the external world.}
\vspace{-0.8em}
\centering
\begin{tabular}{c|c|c|c}
  \toprule
  & Representative work & Simulation focus & Role of world model \\
  \hline
  \multirow{10}{*}{Mirroring Real-world Society} 
  & Generative Agents~\cite{park2023generative} & Daily social life & Stylized facts \\
  & AI Town~\cite{park2023generative} & Sandbox community & Stylized facts \\
  & S3~\cite{gao2023s} & Social networks & Predictions \\
  & Papachristou \textit{et al.}~\cite{papachristou2024network} & Network formation & Stylized facts \\
  & Xu \textit{et al.}~\cite{xu2023exploring} & Social games & Strategic patterns \\
  & EconAgent~\cite{li2024econagent} & Macroeconomics & Stylized facts \\
  & SRAP-Agent~\cite{ji2024srap} & Resource allocation & Stylized facts \\
  & Project Sid~\cite{al2024project} & Collective rules & Stylized facts \\
  & OASIS~\cite{yangoasis} & Social media  & Stylized facts \\
  & GenSim~\cite{tanggensim} & Social media & Stylized facts \\
& YuLan-OneSim~\cite{wang2025yulan} & General platform & Stylized facts \\
  & AgentSociety~\cite{piao2025agentsociety} & General platform & Stylized facts \& predictions \\
  & SocioVerse~\cite{zhang2025socioverse} & General platform & Stylized facts \& predictions \\

  \hline
  \multirow{4}{*}{Understanding the External World} 
  & Agent-Pro~\cite{zhang2024agent} & Interactive games & Belief formation \\
  & Zhang \textit{et al.}~\cite{zhang2023exploring} & Collaboration tasks & Reflection \& debate \\
  & GovSim~\cite{piatti2024cooperate} & Resource sustainability & Collective cognition \\
  & AgentGroupChat~\cite{gu2024agent} & Group deliberation & Belief \& memory \\

  \bottomrule
\end{tabular}
\label{tab:simulacra}
\vspace{-1em}
\end{table}

\subsubsection{Building Social Simulacra Mirroring Real-world Society.}

With the rapid development of LLM agents, building realistic social simulation systems has become increasingly feasible. A well-known example is AI Town~\cite{park2023generative}, a sandbox environment composed of generative agents that exhibit believable individual behaviors and, at the group level, produce emergent dynamics resembling those in real communities. These systems illustrate how social simulacra can serve as explicit world models, providing environments where societal intelligence—collective sensing, reasoning, and coordination—can be observed and studied.

In social networks, S3~\cite{gao2023s} shows that LLM agents can reproduce realistic patterns of information diffusion, capturing the dynamics of public events as they unfold.  Papachristou \textit{et al.}~\cite{papachristou2024network} further highlight the spontaneous formation of network structures among agents, mirroring the self-organization of human societies. Such works reveal how simulacra embody the adaptive and communicative aspects of societal intelligence in digital form. Beyond networks, LLM agents have demonstrated the capacity to model higher-order reasoning in strategic interactions. Xu \textit{et al.}~\cite{xu2023exploring} show that agents in social deduction games like Werewolf display strategic behaviors such as deception and confrontation, reflecting cognitive and competitive dimensions of societal intelligence. In economics and resource allocation, LLM-based agents enable bottom-up modeling of macro-level outcomes from micro-level reasoning. EconAgent~\cite{li2024econagent} reproduces macroeconomic trends emerging from individual decision-making, SRAP-Agent~\cite{ji2024srap} evaluates policy effects in resource allocation, and Project Sid~\cite{al2024project} explores collective responses to taxation rules. These examples illustrate how societal intelligence can manifest as aggregate patterns in economic systems.

More recently, research has shifted toward large-scale platforms that generalize across multiple domains. Among them, \textit{AgentSociety}~\cite{piao2025agentsociety} stands out as the most advanced effort to date, providing a large-scale and versatile environment for examining polarization, policy interventions, and other phenomena central to societal intelligence. Building on this direction, platforms such as GenSim~\cite{tanggensim}, YuLan-OneSim~\cite{wang2025yulan}, OASIS~\cite{yangoasis}, and SocioVerse~\cite{zhang2025socioverse} further extend the vision by integrating diverse social contexts and scaling simulations to thousands or even millions of agents.

Together, these studies demonstrate that LLM-driven social simulacra can serve as explicit world models of human societies, enabling systematic investigations of societal intelligence across social, strategic, and economic domains.

\subsubsection{Agent's Understanding of External World in Social Simulacra}

Beyond mirroring societies at the macro level, social simulacra also enable the study of how agents form internal representations of their environment. Through interaction, LLM agents accumulate experiences, store them as memory, and transform them into implicit world models. These models provide the cognitive substrate for societal intelligence, allowing agents not only to recall past interactions but also to reason about other agents and the broader environment when making decisions~\cite{zhang2024survey}.

Several works demonstrate how implicit world models emerge in practice. Agent-Pro~\cite{zhang2024agent} transforms interaction histories into structured \textit{beliefs}, which then guide subsequent decision-making and strategy updates.  These beliefs reflect an agent’s understanding of others and connect directly to Theory of Mind capabilities discussed in Section~\ref{world_knowledge}. Zhang \textit{et al.}~\cite{zhang2023exploring} further extend this direction by introducing reflection and debate mechanisms from social psychology to improve collaboration in multi-agent tasks.

At the collective level, GovSim~\cite{piatti2024cooperate} investigates whether sustainable cooperation can arise in a society of LLM agents. In this setup, each agent gathers information about shared resources and the behavioral strategies of peers through dialogue, forming higher-level insights about the external environment.  These insights amount to implicit representations of the world model that underpin group-level societal intelligence. Another application is Interactive Group Chat~\cite{gu2024agent}, which explores human-like deliberation across scenarios such as inheritance disputes and court debates.  Here, agents leverage memory and reasoning to produce strategies and social dynamics that closely resemble real human interactions.

\subsection{Functions of World Models}
At their core, world models are designed to receive external commands or actions and model the dynamic state transitions of an environment. Their functionalities can be broadly categorized into two main roles: acting as cloud-based environments and serving as edge-side agent brains. Cloud-based world models typically manifest as video generation systems, synthesizing large volumes of high-quality video data driven by text or action trajectories. This generated data can function as a data engine, augmenting real-world data for training policy models (e.g., VLA and VLN models). Furthermore, cloud-based world models can act as environments within reinforcement learning, interacting with agents to facilitate evolutionary learning in a virtual setting. This capability significantly reduces the cost and risks associated with real-world interactions, proving especially critical in domains like autonomous driving. Additionally, cloud-based world models can serve as policy evaluators, outputting observation sequences through interaction with policy models, thereby enabling the assessment of policy model performance. In contrast, world models acting as edge-side agent brains often do not require low-level visual generation. Instead, they can compress world states within a latent space; for example, V-JEPA 2~\cite{assran2025v} trains a latent space world model that enables on-device action planning through model predictive control. Alternatively, a two-stage approach~\cite{hu2024video} can be employed, where the world model first processes visual observations, which are then converted into executable actions.

\section{Open Problems and Future Directions } \label{sec:discussion}
The recent advance of hyper-realistic generative AI has brought a lot of attention to development of the world model, with particular focus on the multi-modal big models like Sora~\cite{sora2024}. Despite the rapid innovation, there are also a lot of important open problems that remain to be solved.

\subsection{Physical Rules and Counterfactual Simulation}

A key objective of world models is to capture the causal structure of their environments -- especially the underlying physical rules -- so they can reason about counterfactuals beyond the data distribution~\cite{pearl2009causal}. This capacity is crucial for handling rare, mission-critical events (e.g., autonomous-driving corner cases~\cite{feng2023dense}) and for narrowing sim-to-real gaps. Recent progress raises the question of whether large-scale, purely data-driven generative models can acquire such rules from raw visual data alone. While transformer- and diffusion-based video generators such as Sora~\cite{sora2024} produce strikingly realistic sequences, studies reveal persistent physical-law failures -- e.g., inaccurate gravity, fluid, or thermal dynamics~\cite{wang2023newton}.

Hybrid approaches that explicitly embed physics are emerging as promising alternatives. Genesis\cite{Genesis} illustrates this direction by unifying fast, photo-realistic rendering with a re-engineered universal physics core, allowing language-conditioned data generation grounded in first-principles simulation. PhysGen\cite{liu2024physgen} takes a similar stance at the image-to-video level: it couples a rigid-body simulator with a diffusion refiner, enabling controllable, physically plausible motion from a single image. Complementarily, soft-constraint hybrids enforce physics via learning-time priors. physics-informed diffusion introduces PDE-based residual losses that penalize violations of governing equations while preserving generative flexibility~\cite{bastek2025physicsinformed}. Such ``hard+soft'' designs improve controllability and interpretability without sacrificing realism.

Complementary diagnostic work underscores why such hybrids are needed. Kang et al.\cite{kang2024far} show that scaling diffusion video models yields perfect in-distribution fidelity yet breaks down on out-of-distribution or combinatorial tests, indicating “case-based’’ rather than rule-based generalization. Motamed et al.\cite{motamed2025generative} reach a similar conclusion with the Physics-IQ benchmark: current video generators achieve visual realism but largely fail on tasks requiring understanding of optics, fluid dynamics, or magnetism. In parallel, first-principles benchmarks have begun to operationalize ``physical fidelity'' as a measurable axis: T2VPhysBench~\cite{guo2025t2vphysbench} evaluates adherence to core physical laws (including Newtonian mechanics and conservation principles) and documents systematic violations across leading text-to-video systems, while VBench-2.0~\cite{zheng2025vbench} explicitly introduces \emph{Physics} and \emph{Commonsense} as standard evaluation dimensions for video generation.

Taken together, the evidence suggests that data-driven scaling alone is insufficient to recover robust physical laws. Integrating explicit simulators -- or otherwise enforcing physical priors~\cite{shi2022learning} -- remains a promising path toward world models that generalize to unseen counterfactual scenarios while retaining interpretability and transparency.

\subsection{Enriching the Social Dimension}
Simulating the physical elements alone is not sufficient for an advanced world model, since human behavior and social interaction also play a crucial role in many important scenarios~\cite{gao2024large,yuan2025learning,gong2025behavegpt}. For example, the behavior of urban dwellers is particularly important for building world models of the urban environment~\cite{batty2024digital,xu2023urban}. Previous work shows that the human-like commonsense reasoning capabilities of LLMs provide a unique opportunity to simulate realistic human behavior with generative agents~\cite{park2023generative}. However, designing autonomous agents that can simulate realistic and comprehensive human behavior and social interactions remains an open problem. 
Recent studies suggest that theories of human behavior patterns and cognitive processes can inform the design of agentic workflows, which in turn enhance the human behavior simulation capabilities of LLMs~\cite{shao2024beyond,park2023generative}, representing an important direction for future research. In addition, the evaluation of the realism of generated human behavior still largely relies on subjective human assessment, which is challenging to scale up to a large-scale world model. Developing a reliable and scalable evaluation scheme will be another future research direction that can enrich the social dimension of the world model. 

\subsection{Benchmarks}

\begin{table}[t]
\footnotesize
\renewcommand{\arraystretch}{0.9}
\caption{Representative benchmarks for evaluating world models.}
\vspace{-1em}
\centering
\begin{tabular}{
  >{\centering\arraybackslash}p{.18\linewidth}|
  >{\centering\arraybackslash}p{.22\linewidth}|
  >{\centering\arraybackslash}p{.50\linewidth}
}
  \toprule
  \textbf{Category} & \textbf{Benchmark} & \textbf{Scope \& highlights} \\
  \hline
  \multirow{5}{*}{\makecell{Video-centric\\ world simulation}}
  & WorldSimBench~\cite{qin2024worldsimbench}
    & Sandbox/driving/manipulation; links human preference to \emph{action-level consistency} \\
  & WorldScore~\cite{duan2025worldscore}
    & 3{,}000 camera-specified scenes; decomposes \emph{controllability/quality/dynamics}; compares 3D/4D/video generators \\
  & VBench~\cite{huang2024vbench}
    & General T2V/V2V; automatic axes: \emph{temporal consistency}, stability, prompt adherence \\
  & VBench-2.0~\cite{zheng2025vbench}
    & General T2V/V2V; \emph{intrinsic faithfulness}: physics, commonsense, human fidelity, controllability \\
  & T2V-CompBench~\cite{sun2025t2vcompbench}
    & Compositional T2V; tests \emph{binding} of attributes/actions/relations with camera motion \\
  \hline
  \multirow{6}{*}{\makecell{Physical \&\\ spatial reasoning}}
  & PhysBench~\cite{chow2025physbench}
    & 10k video–image–text triplets; probes \emph{properties/relations/dynamics} gaps in VLMs \\
  & UrbanVideo-Bench~\cite{zhao2025urbanvideo}
    & 5.2k urban egocentric clips (16 tasks); \emph{recall, navigation, causal reasoning} \\
  & Physics-IQ~\cite{motamed2025generative}
    & Five domains (solids/fluids/optics/thermo/magnetism); \emph{law adherence vs.\ perceived realism} \\
  & T2VPhysBench~\cite{guo2025t2vphysbench}
    & Text-to-video; first-principles checklist of \emph{12 core laws} \\
  & VideoPhy~\cite{bansal2025videophy}
    & Action-centric prompts; \emph{semantic adherence} \& \emph{physical commonsense}; rule attribution \\
  & Basic Spatial Abilities~\cite{xu2025bsa}
    & Psychometric framing; five spatial skills (perception/relations/orientation/rotation/visualization) \\
  \hline
  \multirow{4}{*}{\makecell{Embodied\\ decision making}}
  & EAI~\cite{li2024embodied}
    & LLM-based agents; module-level (goal/subgoal/action/transition) evaluation; error taxonomy \\
  & EWMBench~\cite{yue2025ewmbench}
    & AgiBotWorld manipulation/navigation; \emph{scene consistency}, \emph{motion correctness}, semantic alignment \\
  & WPE~\cite{quevedo2025evaluating}
    & Policy evaluation in world model vs.\ real/sim; \emph{sequence-/part-level correspondence} under identical actions \\
  & RoboScape~\cite{shang2025roboscape}
    & Physics-informed embodied world model; \emph{policy lift} \& \emph{sim-to-real gap} (data-engine role) \\
  \bottomrule
\end{tabular}
\label{tab:benchmarks}
\vspace{-1em}
\end{table}

Benchmarking world models is both necessary and challenging. Because the community pursues divergent goals—\textit{learning internal representations} vs.\ \textit{predicting future worlds}—with heterogeneous technical approaches (e.g., LLM agents, video diffusion) and wide-ranging application domains (autonomous driving, robotics, social simulation), there is no single canonical task or metric. Nevertheless, several recent efforts illustrate how carefully designed testbeds can expose the specific gaps that prevent current models from becoming reliable world simulators, as summarized in Table~\ref{tab:benchmarks}.

\textbf{Video-centric world simulation.} WorldSimBench links perceptual quality to control by pairing human preferences with action-level consistency across sandbox, driving, and manipulation settings~\cite{qin2024worldsimbench}; WorldScore complements this with a camera-specified protocol that decomposes performance into controllability, visual quality, and dynamics over 3,000 scenes, enabling head-to-head comparisons among 3D/4D and video generators~\cite{duan2025worldscore}. VBench operationalizes largely automatic axes—temporal consistency, subject/background stability, and prompt adherence~\cite{huang2024vbench}—while VBench-2.0 elevates \emph{intrinsic faithfulness} (physics, commonsense, human fidelity, controllability) to separate ``looking real'' from ``acting like a world''~\cite{zheng2025vbench}. Compositional stress tests arrive via T2V-CompBench, which measures binding among attributes, actions, relations, and camera motion using MLLM-, detection-, and tracking-based metrics~\cite{sun2025t2vcompbench}.

\textbf{Physical and spatial reasoning.} Beyond appearance, world models must respect physical law and support spatial competence. PhysBench (10k video–image–text triplets) reveals systematic gaps in object properties, relations, and dynamics for modern VLMs~\cite{chow2025physbench}; UrbanVideo-Bench (5.2k drone clips, 16 task types) diagnoses deficits in recall, navigation, and causal reasoning in long egocentric streams~\cite{zhao2025urbanvideo}. Law-centric suites sharpen the picture: Physics-IQ evaluates five domains (solid/fluid mechanics, optics, thermodynamics, magnetism), finding physical understanding largely decoupled from perceived realism~\cite{motamed2025generative}; T2VPhysBench provides a first-principles checklist of 12 core laws for text-to-video systems~\cite{guo2025t2vphysbench}. On the generative side, VideoPhy quantifies \emph{semantic adherence} and \emph{physical commonsense} under action-centric prompts, attributing errors to concrete rules (e.g., support, inertia, continuity)~\cite{bansal2025videophy}. A complementary psychometric framing anchors five Basic Spatial Abilities—perception, relations, orientation, mental rotation, visualization—revealing geometry/rotation weaknesses across 13 VLMs and supplying calibrated tasks for progress tracking~\cite{xu2025bsa}.

\textbf{Embodied decision making.} When world models are embedded in control loops, aggregate success rates obscure process failures. The Embodied Agent Interface (EAI) standardizes four LLM-based modules—goal interpretation, subgoal decomposition, action sequencing, transition modeling—and reports fine-grained error taxonomies (e.g., hallucination, affordance, planning)~\cite{li2024embodied}. EWMBench tests embodied world models on scene consistency, motion correctness (physics-/task-consistent trajectories), and semantic alignment using AgiBot World data, explicitly tying video plausibility to action preconditions and affordances~\cite{yue2025ewmbench}. A role-centered view evaluates models as \emph{environments}: WPE compares rollouts under identical action sequences in the model versus real videos/simulators, reporting sequence- and part-level correspondences~\cite{quevedo2025evaluating}; physics-informed world models such as RoboScape assess the \emph{data-engine} role by measuring policy lift and sim-to-real gap when training on model-generated experience~\cite{shang2025roboscape}.

Despite these advances, benchmarking world models remains an open challenge. Future work should focus on building more diverse and realistic benchmarks to rigorously test generalization capabilities. Additionally, standardizing evaluation protocols will be key to improving comparability and robustness assessments across environments.

\subsection{Bridging Simulation and Reality with Embodied Intelligence}
The world model has long been envisioned as a critical step towards developing embodied intelligence~\cite{savva2019habitat}. It can serve as a powerful simulator that creates comprehensive elements of the environment and models realistic relationships between them. Such an environment can facilitate embodied agents to learn through interaction with a simulated environment, reducing the need for supervision data. To achieve this goal, improving the multi-modal, multi-task, and 3D capacities of generative AI models has become an important research problem for developing general world models for embodied agents. Moreover, closing the simulation-to-reality gaps~\cite{hofer2021sim2real} has been a long-standing research problem for embodied environment simulators, and it is therefore important to transfer the trained embodied intelligence from the simulation environment to the physical world. Collecting more fine-grained sensory data is also a critical step toward this goal, which can be facilitated through the interface of embodied agents. Therefore, an interesting future research direction is to create self-reinforcing loops to harness the synergy power of generative world models and embodied agents.

\subsection{Simulation Efficiency}

Ensuring high simulation efficiency of world models is important for many applications. For example, number of frames per second is a key metric for high quality for learning sophisticated drone manipulating AIs. The popular transformer architecture of most big generative AIs poses a huge challenge for high-speed simulation because its autoregressive nature can only generate one token at a time. Several strategies are proposed to accelerate the inference of large generative models, such as incorporating big and small generative models~\cite{shang2024defint} and distilling big models~\cite{shao2024beyond}. More holistic solutions include building a simulation platform that optimally schedule LLM requests~\cite{yan2024opencity}.
High computation cost is also a problem for classic physics simulators when they are tasked to simulate large and complex systems. Previous research finds deep learning models like graph neural networks can be used to efficiently approximate physical systems~\cite{sanchez2020learning}. Therefore, an important research direction will be to explore the synergy between smaller deep learning models and big generative AI models. Additionally, the overall improvement from underlying hardware to programming platform and AI models is also needed to achieve substantial speedup.  

\subsection{Ethical and Safety Concerns}

\textbf{Data Privacy.} The recent trend of building world models with big generative AIs raises significant concerns of privacy risk, largely due to the massive and often opaque training data~\cite{yao2024survey}. Extensive research effort is devoted to assessing the risk of inferring private information with big generative AIs like LLM~\cite{li2024llm}, which could be especially sensitive in the context of video generation models. To be compliant with privacy regulations like GDPR~\cite{tamburri2020design}, it is important to improve the transparency of the life cycle of generative AIs, helping the public understand how data is collected, stored, and used in these AI models.

\textbf{Simulating Unsafe Scenario.} The incredibly intelligent power of generative AIs makes safeguarding their access a paramount task. Previous studies on LLMs found they can be misled to generate unsafe content with adversarial prompting~\cite{kumar2023certifying,inan2023llama}. The risk of unsafe use of world models can be even larger. Adversarial users might leverage such techniques to simulate harmful scenarios, reducing the cost of planning illegal and unethical activities. Therefore, an important future research direction is to safeguard the usage of world models.

\textbf{Accountability.} The ability to generate hyper-realistic text, images, and videos has caused severe social problems of spreading misinformation and disinformation. For example, the emergence of deepfake technology gives rise to large-scale misuses that have widespread negative effects on social, economic, and political systems~\cite{westerlund2019emergence}. Thus, detecting AI-generated content has been a key research problem in addressing these risks~\cite{rana2022deepfake}. However, this problem is becoming increasingly challenging due to the advance of generative AIs, and it will be even more difficult with the arrival of a world model that can generate consistent, multi-dimensional output. Technology like watermarking could help improve the accountability of world model usage~\cite{dathathri2024scalable}. More research attention, as well as legal solutions, are needed to improve the accountability of world model usage.

\section{Conclusion}\label{sec::conclusion}

Understanding the world and predicting the future have been long-standing objectives for scientists developing artificial generative intelligence, underscoring the significance of constructing world models across various domains. This paper presents the first comprehensive survey of world models that systematically explores their two primary functionalities: \textit{implicit representations} and \textit{future predictions} of the external world. We provide an extensive summary of existing research on these core functions, with particular emphasis on world models in decision-making, world knowledge learned by models, world models as video generation, and world models as embodied environments. Additionally, we review progress in key applications of world models, including generative games, robotics, autonomous driving, and social simulacra. Finally, recognizing the unresolved challenges in this rapidly evolving field, we highlight open problems and propose promising research directions with the hope of stimulating further investigation in this burgeoning area.

\bibliographystyle{plain}
\bibliography{bibliography}

\appendix
\counterwithin*{table}{section} %
\renewcommand{\thetable}{S\arabic{table}} 
\counterwithin*{figure}{section} %
\renewcommand{\thefigure}{S\arabic{figure}} 
\section{Related survey}
\begin{table}[h]
\centering
\caption{Comparison with existing surveys. This paper focuses on a comprehensive overview of the systematic definition and the capabilities of world models.}
\begin{tabular}{c|ccc}
\toprule
Survey & Venue and Year & Main Focus & Deficiency \\
\midrule
\cite{zhu2024sora} & Arxiv, 2024 & General world model & Limited to discussion on applications \\
\cite{mai2024efficient} & Arxiv, 2024 & Efficient multimodal models & Limited to discussion on techniques \\
\cite{cho2024sora} & Arxiv, 2024 & Text-to-video generation & Limited scope \\
\cite{guan2024world} & IEEE T-IV, 2024 & Autonomous driving & Limited scope \\
\cite{li2024data} & Arxiv, 2024 & Autonomous driving & Limited scope \\
\cite{yan2024forging} & Arxiv, 2024 & Autonomous driving & Limited scope \\
\bottomrule
\end{tabular}
\label{tab:survey_compare}
\end{table}

\section{Figures and tables}
\begin{figure}[h]
    \centering
    \includegraphics[width=0.9\linewidth]{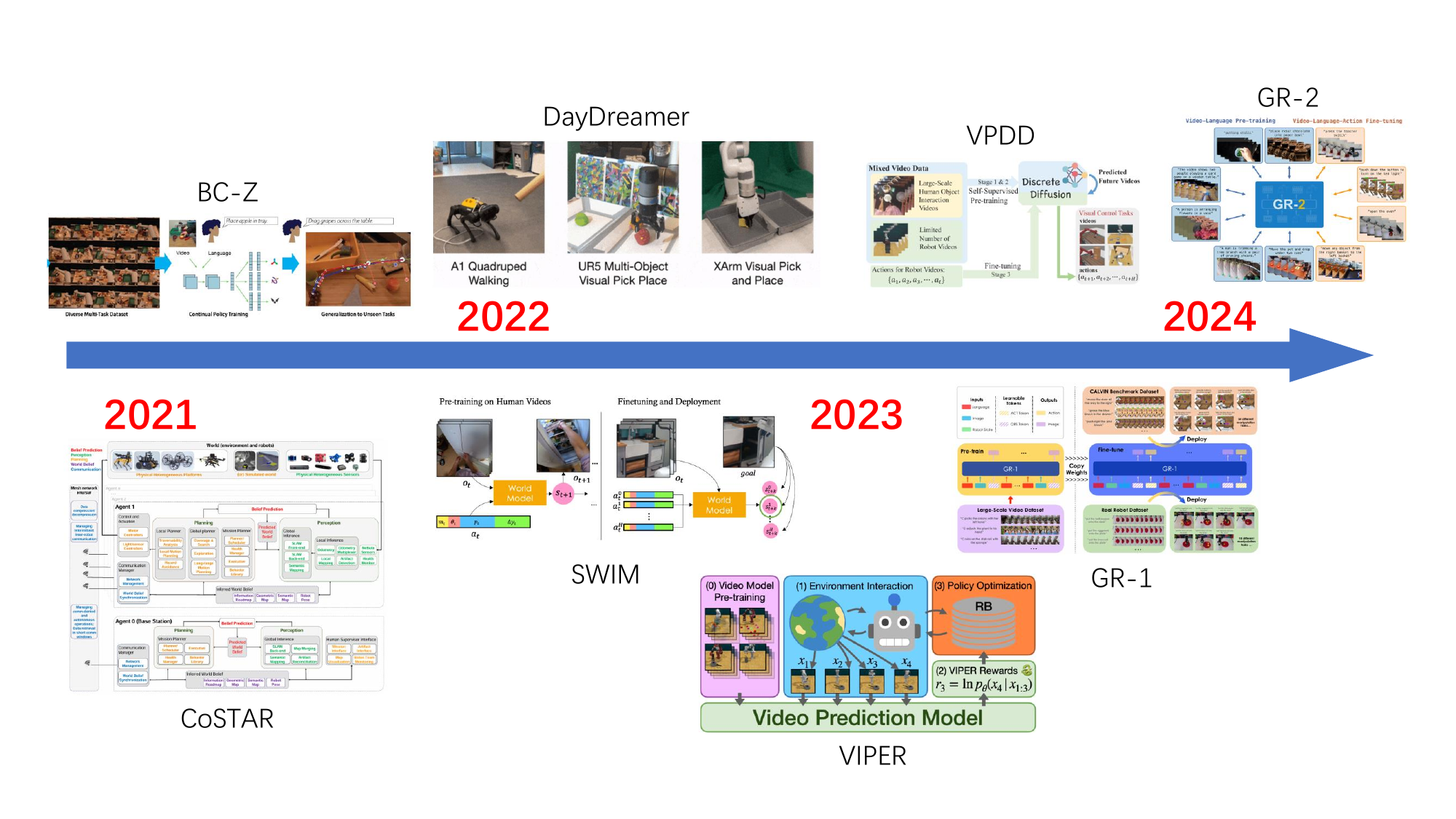}
    \caption{The development of the robotic world model.}
    \label{fig:robot_wm}
\end{figure}

\begin{figure}[h]
    \centering
    \includegraphics[width=0.6\linewidth]{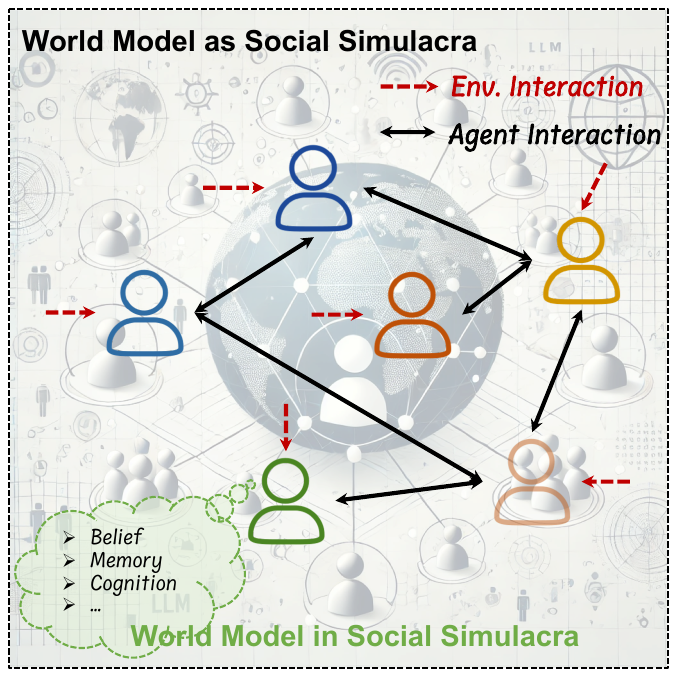}
    \caption{World model and social simulacra.}
    \label{fig:simulacra}
\end{figure}

\section{Update History}
\begin{itemize}
    \item[-] 2025.09.09: published version for ACM Computing Survey.
    \item[-] 2025.11.10: rewrite Sec.~\ref{sec:background} (History and Current Development), summarize the roadmap of world models in the deep learning era (Fig.~\ref{fig:intro}), reorganize application domains in Sec.~\ref{sec:application}, update recent papers.
\end{itemize}

\end{document}